\documentclass{article}
\usepackage{graphicx} 
\usepackage[utf8]{inputenc}
\usepackage{amsmath}
\usepackage{amssymb}
\usepackage{amsthm}
\usepackage{dsfont}
\usepackage[a-1b]{pdfx} 
\usepackage{hyperref}
\usepackage{csquotes}
\usepackage{pdflscape}
\usepackage[a4paper, total={6in, 8.5in}]{geometry}
\hypersetup{colorlinks,linkcolor={black},citecolor={blue},urlcolor={black}}  
\usepackage[english]{babel}
\usepackage{comment}
\usepackage{bbm}
\usepackage{float}
\usepackage{multirow}
\usepackage{etoolbox}
\usepackage{makecell}

\usepackage{algorithm}%
\usepackage{algorithmicx}%
\usepackage{algpseudocode}%
\usepackage{listings}%
\usepackage{chngcntr}

\usepackage[round]{natbib}
\usepackage[toc,page]{appendix}

\usepackage{authblk}

\newtheorem{theorem}{Theorem}[section]

\newtheorem{lemma}[theorem]{Lemma}
\newtheorem{proposition}[theorem]{Proposition}
\newtheorem{assumption}[theorem]{Assumption}

\newtheoremstyle{mystyle}
  {\topsep}
  {\topsep}
  {\normalfont}
  {}
  {\bfseries}
  {.}
  {.5em}
  {}

\theoremstyle{mystyle}
\newtheorem{remark}[theorem]{Remark}

\newcommand{\pr}{\mathbb{P}}

\newcommand{\Ev}{\mathbb{E}}
\newcommand{\ind}{\mathds{1}}

\newcommand{\real}{\mathbb{R}}

\newcommand{\setc}{\mathcal{C}}

\title{Conformal online model aggregation}
\author[1]{Matteo Gasparin}
\author[2,3]{Aaditya Ramdas}
\affil[3]{Machine Learning Department, Carnegie Mellon University}
\affil[2]{Department of Statistics and Data Science, Carnegie Mellon University}
\affil[1]{Department of Statistical Sciences, University of Padova}
\date{}

\begin{document}
\date{\today}
\maketitle

\begin{abstract}
Conformal prediction equips machine learning models with a reasonable notion of uncertainty quantification without making strong distributional assumptions. It wraps around any prediction model and converts point predictions into set predictions with a predefined marginal coverage guarantee. However, conformal prediction only works if we fix the underlying machine learning model in advance. A relatively unaddressed issue in conformal prediction is that of model selection and/or aggregation: given a set of prediction models, which one should we conformalize? This paper suggests that instead of performing model \emph{selection}, it can be prudent and practical to perform \emph{conformal set aggregation} in an online, adaptive fashion. We propose a wrapper that takes in several conformal prediction sets (themselves wrapped around black-box prediction models), and outputs a single adaptively-combined prediction set.
Our method, called conformal online model aggregation (COMA), is based on combining the prediction sets from several algorithms by weighted voting, and can be thought of as a sort of \emph{online stacking of the underlying conformal sets}. As long as the input sets have (distribution-free) coverage guarantees, COMA retains coverage guarantees, under a negative correlation assumption between errors and weights. We verify that the assumption holds empirically in all settings considered. COMA is well-suited for decentralized or distributed settings, where different users may have different models, and are only willing to share their prediction sets for a new test point in a black-box fashion. As we demonstrate, it is also well-suited to settings with distribution drift and shift, where model selection can be imprudent.
\end{abstract}

\section{Introduction}
\label{sec:intro}
Conformal prediction has emerged as a significant approach within the field of machine learning, offering a reliable wrapper for uncertainty quantification. Unlike traditional methods that rely on strict distributional assumptions, conformal prediction provides a flexible framework by transforming point predictions obtained with any black-box model into prediction sets with valid finite-sample marginal coverage. Initially introduced by \citet{saunders1999transduction}, conformal prediction has gained increasing prominence and has been extended beyond the iid (or exchangeable) data setting; some examples are \citet{tibshirani2019, gibbs2021, barber2023}.

A significant challenge in the application of conformal prediction arises in the context of model selection or aggregation. The conventional approach involves fixing the underlying machine learning (or prediction) model in advance and then obtaining an interval with a prespecified level of marginal coverage. Since different models yield sets with the same theoretical coverage guarantees (in iid settings), the natural goal is to select the best of a collection of models. In online settings with distribution drift, it appears imprudent to commit to one model (model selection), and hence a better idea is to aggregate the models in an adaptive manner (model aggregation).

\begin{figure}
    \centering
    \includegraphics[width=0.88\linewidth]{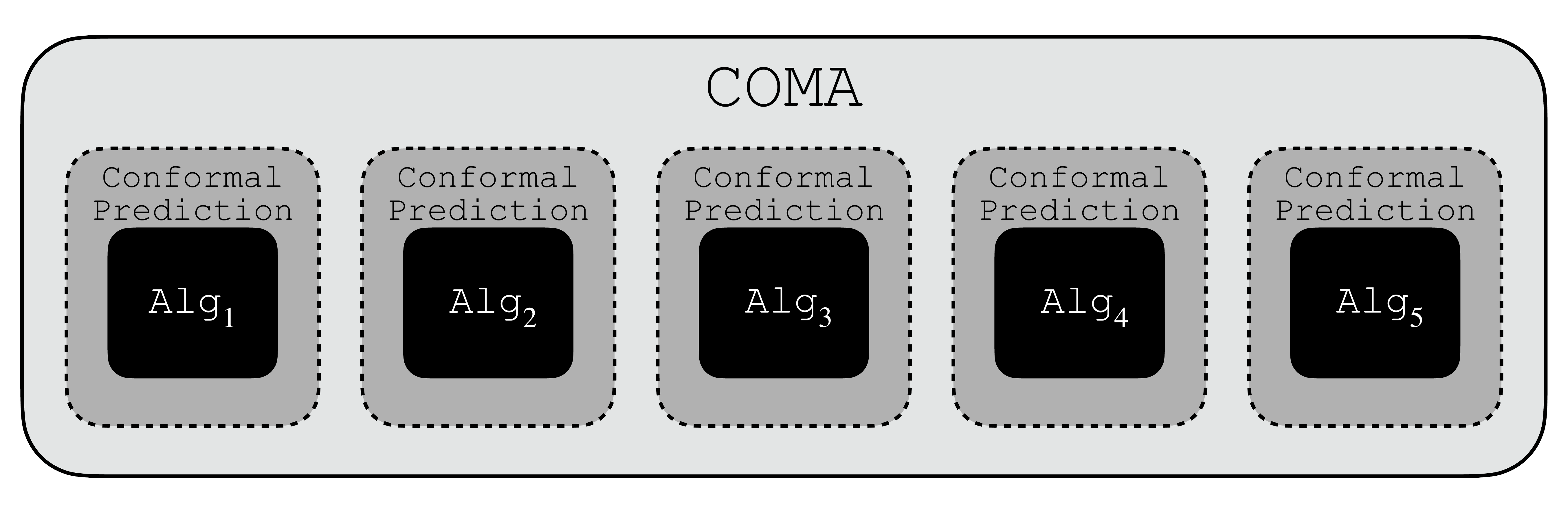}
    \caption{Graphical summary of the COMA method. Conformal prediction wraps around the individual prediction algorithms, while COMA operates as a higher-level wrapper that aggregates these sets}
    \label{fig:coma_il}
\end{figure}

We propose a method that combines the prediction sets produced by different prediction models through a voting mechanism. It acts as a wrapper over the sets produced by conformal prediction, as illustrated in Figure~\ref{fig:coma_il}, and can be seen as a stacking of conformal sets, where the weights assigned to each model in the aggregation process are dynamically adjusted over time to reflect their past performance. The performance of the various models is assessed based on the efficiency of the sets, such as cardinality for a discrete set or Lebesgue measure for a continuous set. Specifically, models consistently producing smaller intervals will be favored over others. We examine both iid and online scenarios with potential distributional shift, and show that our proposed solution, called COMA, performs effectively into both frameworks. In particular, in the latter case, it is used in conjunction with the adaptive conformal approaches proposed by \cite{gibbs2021} and \cite{angelopoulos2023pid,angelopoulos2024decaying}. Our solution provides a regret bound on the size of the aggregated prediction set and maintains valid coverage under a negative correlation assumption between model errors and aggregation weights, an assumption that is empirically verified in our experiments. Moreover, since the approach only needs the underlying conformal prediction sets, it is particularly well-suited for distributed settings where users are only willing to share their prediction sets.

\subsection{Related work}
Conformal prediction was introduced by Vovk and collaborators, with the first book on the topic dating back to \citet{vovk2005}. Several other influential papers in the ``offline'' setting include \citet{papadopoulos2002}, \citet{lei2018}, \citet{romano2019}, \citet{romano2020} and \citet{barber2021}. 
All these approaches provide reliable uncertainty quantification for a single black-box algorithm and are grounded on the assumption that observations are iid or at least exchangeable. \cite{cherubin2019} and \cite{solari2022} introduced a technique to merge conformal prediction intervals using a simple majority voting scheme. This method was further refined by \cite{gasparin2024}, who added a weighting system and the use of randomization. Their proposal can be viewed as inverting method based on the combination of arbitrarily dependent p-values. Although there are other methods in the literature to combine p-values \citep{vovk2022, toccaceli2017, vovk2015}, the majority vote does not require the need to have direct access to the conformal p-values for each value, but it only requires the sets obtained by the different agents. \citet{humbert2023} propose a method for aggregating the results of various agents within a conformal prediction framework. This method specifically addresses the context of federated learning, where privacy concerns restrict agent operations. The problem of model selection, within the realm of conformal prediction, is studied by \citet{yang2021} and \citet{fan2023}. Our work is distinct from theirs, as it operates within an online setting and in particular non-iid scenarios. Furthermore, rather than performing model selection, we focus on model aggregation.

In recent years, numerous works have aimed to extend conformal prediction beyond the context of iid or exchangeable data \citep{chernozhukov2018exact, tibshirani2019, podkopaev2021distribution, dunn2023distribution, barber2023}. In particular, some of these studies have specialized in the case where data arrive in an online fashion and can exhibit a continuous shift in the distribution. One of the pioneering works is \citet{gibbs2021}, where the main idea is to adapt the error level $\alpha$ over time in order to achieve the desired error rate. Several works extend this method in different ways; examples include \cite{zaffran2022, gibbs2023, angelopoulos2023pid, angelopoulos2024decaying}. 

Lastly, outside the conformal framework, the problem of combining expert advice is well-studied in online learning. It is customary that at each iteration a weight is assigned to the agent, which is subsequently updated according to its performance. One of the most widely employed methods is the Exponential Weighted (\textsc{ew}) algorithm \citep{littlestone1994,freund1997}. While we are inspired by such ideas, their usual goals of regret minimization for bounded adversarial losses are not directly well suited to our setting (minimizing the size of the merged prediction sets), but we will show how to adapt particular variants for our purposes.  
Our work may be the first to merge such online learning techniques with conformal prediction.

\subsection{Summary of contributions}
We now concisely highlight the key contributions of the work, anticipating some of results presented later:
\begin{itemize}
    \item In Sections~\ref{sec:problem_setup} and~\ref{sec:weighted}, we introduce a \emph{wrapper method} which takes as input the prediction sets produced by $K$ predictive algorithms and outputs a single, adaptively combined prediction set. The proposed approach aggregates the individual sets through a weighted voting scheme, where the weights are dynamically updated according to the performance of each algorithm. We analyze the theoretical properties of the method and show in Theorem~\ref{prop:cov} that if the base prediction sets have miscoverage at most $\alpha$, then the combined set achieves miscoverage at most $2\alpha$ under a negative dependence assumption between the errors and the weights. The assumption is verified empirically in our experiments (Section~\ref{sec:dyn_real}); moreover, when this assumption is violated, the coverage degrades smoothly as a function of the degree of violation, as stated in Remark~\ref{rem:ass}.
    \item In Section~\ref{sec:dynamic}, we propose a weighting scheme that updates the weights assigned to each algorithm based on their efficiency (i.e., the measure of the sets). The procedure builds upon a modified version of the Exponentially Weighted algorithm proposed by \citet{derooij2014}. Theorem~\ref{th:bound} shows that the weighting mechanism guarantees a regret bound on the cumulative loss with respect to the best-performing algorithm, even in the presence of unbounded losses.
    \item In Section~\ref{sec:dyn_aci}, we extend our method to scenarios with potential distribution shift. In particular, the proposed COMA framework can be  integrated with the adaptive conformal inference approaches of \citet{gibbs2021} and \citet{angelopoulos2023pid,angelopoulos2024decaying}. We investigate two complementary settings. In the first, the agents (i.e., base algorithms) operate in a decentralized fashion, and at each step our method only combines their individual prediction sets. In this case, {COMA} guarantees a long-run coverage of at least $1 - 2\alpha$, under a negative dependence assumption between the errors and the adaptive weights (Proposition~\ref{prop:long_run}). In the second scenario, the agents act centrally, and the method achieves the same coverage guarantees as adaptive conformal inference by \citet{gibbs2021}.
\end{itemize}

\paragraph{Paper outline.}
The remainder of the paper is organized as follows. Sections~\ref{sec:problem_setup} and~\ref{sec:weighted} formulate the problem and describe the proposed approach, along with its theoretical properties. Section~\ref{sec:dynamic} focuses on a specific weighting scheme based on a modification of the \textsc{ew} algorithm, which enables the establishment of an upper bound on the observed loss in relation to the best-performing algorithm. Section~\ref{sec:dyn_aci} extends the method to scenarios with potential distributional shifts and analyzes its theoretical properties in this context. Section~\ref{sec:dyn_real} demonstrates the practical effectiveness of the proposed methods in both the iid and non-iid settings, before we conclude in Section~\ref{sec:conclusion}.  

\section{Problem setup}
\label{sec:problem_setup}
We focus our attention on an online setting in which data arrive sequentially over time. A plausible scenario arises when dealing with a continuous flow of data; in symbols, for each iteration $t = 1,2,...,T$, a covariate-response pair $z^{(t)} = (x^{(t)}, y^{(t)}) \in \mathcal{X} \times \mathcal{Y}$ representing an instance from the random vector $Z^{(t)} = (X^{(t)}, Y^{(t)})$ is observed. Technically, we first observe $X^{(t)}$, then output a prediction set $\mathcal{C}^{(t)}$ that aims to contain $Y^{(t)}$, and we then observe $Y^{(t)}$. 

We assume access to $K$ different prediction algorithms (or \emph{models} or \emph{experts}) used to predict $Y^{(t)}$ based on covariates $X^{(t)}$. As a result, in each iteration, for any prespecified error tolerance $\alpha \in (0,1)$, we can derive $K$ distinct conformal prediction sets (one for each model) $\mathcal{C}_1^{(t)}, \dots, \mathcal{C}_K^{(t)}$ based on the information from $(\{z^{(i)}\}_{i=1}^{t-1}, x^{(t)})$. Actually, one can choose to use only a subset of $\{z^{(i)}\}_{i=1}^{t-1}$ to produce the sets due to computational issues or distribution shifts. If the data are iid (or exchangeable), conformal prediction sets are known to have valid marginal coverage:  
\begin{equation}
\label{eq:coverage}
    \pr \left(Y^{(t)} \in \setc_k^{(t)}\right) \geq 1-\alpha, \quad k=1,\dots,K,
\end{equation}
where the probability is taken with respect to all sources of randomness (the new data point, the training points, possibly the algorithm). We refer the reader to Appendix~\ref{sec:app_ci} for an introduction to conformal inference. 

The main question that arises is how to efficiently merge or select among the different conformal prediction sets. This issue is particularly relevant in distributed settings, where the aggregator has access only to the $K$ sets, but not to the underlying data or individual model predictions. A key challenge is that these sets are not independent, since they are all constructed from the same underlying data.

We build on an approach recently proposed by \citet{gasparin2024}, which involves a weighted majority vote. Let $w^{(t)} = (w^{(t)}_1, \dots, w^{(t)}_K)$ be a set of weights such that 
\begin{gather*}
    w^{(t)}_k \in [0, 1], \quad k = 1, \dots, K, \text{ and }
    \sum_{k=1}^K w^{(t)}_k = 1,
\end{gather*}
where the weight $w_k^{(t)}$ represents the importance of the prediction algorithm $k$ at time $t$ in the voting procedure. We can define the \emph{weighted majority vote set} as
\begin{equation}
\label{eq:cm}
    \setc_M^{(t)} := \left\{y \in \mathcal{Y}: \sum_{k=1}^K w_k^{(t)} \ind\left\{y \in \setc_k^{(t)}\right\} > \frac{1 + u^{(t)}}{2} \right\},
\end{equation}
where $u^{(t)}$ is a realization from $U^{(t)}$ that is a random variable uniformly distributed in the interval $[0,1]$ independent of all the data. If the weights are chosen a priori (independent of the data), \citet{gasparin2024} prove that if the starting intervals are marginally valid (they respect~\eqref{eq:coverage}) then the coverage of $\setc_M^{(t)}$ is at least $1-2\alpha$. The factor 2 is provably unavoidable and arises from the arbitrary nature of the dependence between sets in this scenario. Moreover, as pointed out in \citet{gasparin2024}, the majority vote set can be computed in $O(K\log K)$ time for real-valued sets. This is due to the need of sorting the interval endpoints and making a single pass over them. In the discrete scenario, the process is straightforward and requires only the counting of intervals that include a specific label.

{It is worth noting that the method requires only the $K$ input sets to construct $\setc_M^{(t)}$. This feature is particularly advantageous in decentralized settings, where the aggregator, at each time step, has access solely to the sets themselves rather than to the conformal p-values, the scores, or the raw outputs of the individual models.}

There are three hurdles to applying these ideas directly:
\begin{itemize}
    \item First, when weights are based on past data, the weights themselves become random variables, so their theory does not immediately apply. This limitation is addressed in Section~\ref{sec:weighted}, where we extend the theoretical results to the case where the weights themselves are random variables.
    \item The second question is how one should update the weights in a theoretically sound way that performs well practically. In Section~\ref{sec:dynamic}, we present a method for updating the weights based on the performance of the different models. In particular, algorithms that consistently produce smaller sets are favored over others. 
    \item Finally, when the data distribution shifts over time, conformal prediction no longer guarantees marginal validity. In Section~\ref{sec:dyn_aci}, we extend the method and use it in conjunction with adaptive conformal prediction techniques to handle such settings.
\end{itemize}

\section{Weighted majority with data-dependent weights}
\label{sec:weighted}
Suppose that the weight vector $w^{(t)}$ is a realization of a random vector $W^{(t)}$ that maps the first $t-1$ observations of $\{Z^{(i)}\}_{i=1}^{t-1}$ into the $(K-1)$-dimensional simplex $\Delta_{K-1}:=\{w \in [0,1]^K: \mathbf{1}^\top w = 1\}$, or more succinctly, $W^{(t)}: (\mathcal{X}\times\mathcal{Y})^{t-1} \to \Delta_{K-1}$. In the following, we define the miscoverage indicator for the $k$-th set at time $t$ as $\phi_k^{(t)}:=\ind\{y^{(t)} \notin \setc_k^{(t)}\}$ and $[K] := \{1,\dots,K\}$. In the following, we will make the assumptions:
\begin{assumption}\label{ass:1}
    Marginal coverage and negative total elementwise correlation between $\phi^{(t)}$, $W^{(t)}$: 
        \begin{equation}\label{eq:ass-neg-tot-corr}
        \Ev\left[\phi_{k}^{(t)} \right] \leq \alpha ~ \mathrm{ for } ~ k \in [K], ~ \mathrm{ and } ~
        \Ev\left[\sum_{k=1}^K\left(\phi_k^{(t)} - \Ev[\phi_k^{(t)}]\right)\left(W_k^{(t)} - \Ev[W_k^{(t)}]\right)\right] \leq 0.
        \end{equation}
\end{assumption}

While marginal coverage is guaranteed if data are iid or at least exchangeable; $\phi^{(t)}$, $W^{(t)}$ could be negatively correlated if, for example, predictors that were assigned high weights (based on previous data) are less likely to make mistakes.
    
\begin{theorem}\label{prop:cov}
    Let $\setc_1^{(t)}, \dots, \setc_K^{(t)}$ be $K \geq 2$ different conformal prediction sets for $Y^{(t)}$, let $W^{(t)}$ be a random vector in $\Delta_{K-1}$ depending only on $\{Z^{(i)}\}_{i=1}^{t-1}$, and consider the weighted majority vote set $\setc_M^{(t)}$ in~\eqref{eq:cm}. It always holds that $\pr(Y^{(t)} \notin \setc_M^{(t)}) \leq 2 \Ev[\max_{k \in [K]} (\phi_1^{(t)}, \dots, \phi^{(t)}_K)]$.
    If Assumption~\ref{ass:1} holds, then we further have $\pr(Y^{(t)} \notin \setc_M^{(t)}) \leq 2\alpha$.  {This implies that with probability at least $1-2\alpha$ the set $\setc_{M}^{(t)}$ is non-empty.}
\end{theorem}
The proof is based on the recently introduced uniformly-randomized Markov inequality \citep{ramdas2023}. 
\begin{proof}
    By direct calculation, we can write
    \[
    \begin{split}
        \pr\left(Y^{(t)} \notin \setc_M^{(t)}\right) &= \pr\left(\sum_{k=1}^K W_k^{(t)} \phi_k^{(t)} \geq \frac{1}{2} - U^{(t)}/2 \right) \stackrel{(i)}{\leq} 2\Ev \left[\sum_{k=1}^K W_k^{(t)} \phi_k^{(t)} \right] 
    \end{split}
    \]
    where $(i)$ is valid due to the Uniformly-randomized Markov Inequality described in \citet{ramdas2023}. 

    The first part of the theorem is a direct consequence of H\"older inequality. Indeed, due to the fact that $||W^{(t)}||_1=1$, we notice that
    \[
    \Ev\left[\sum_{k=1}^K \phi_k^{(t)} W_k^{(t)}\right] \leq \Ev\left[||\phi^{(t)}||_\infty ||W^{(t)}||_1\right] = \Ev\left[||\phi^{(t)}||_\infty\right],
    \]
    where $\phi^{(t)}=(\phi_1^{(t)}, \dots, \phi_K^{(t)})$.

    In the case of negative total elementwise correlation it holds that $\sum_{k=1}^K \Ev[W_k^{(t)}\phi_k^{(t)}] \leq \sum_{k=1}^K \Ev[W_k^{(t)}] \Ev[\phi_k^{(t)}]$ and under marginal coverage, we have that
    \[
    \pr\left(Y^{(t)} \notin \setc_M^{(t)}\right) \leq 2 \sum_{k=1}^K \Ev \left[ W_k^{(t)} \phi_k^{(t)} \right] \leq 2 \sum_{k=1}^K \Ev[W_k^{(t)}]\,\Ev[\phi_k^{(t)}] \leq 2\alpha \sum_{k=1}^K \Ev[W_k^{(t)}] = 2\alpha.
    \]
    Since the target is contained with probability at least $1 - 2\alpha$, it follows that the set is non-empty with at least the same probability.
\end{proof}


\begin{remark}
    The first part of the theorem states, without any assumption, that the probability of miscoverage is bounded by $2 \Ev[\max_{k \in [K]} (\phi_1^{(t)}, \dots, \phi^{(t)}_K)]$. This can translate into a trivial bound (like 1) if there is at least one algorithm that always makes an error, while it can be approximately $2\alpha$ in the case where miscoverage events are strongly correlated and $\Ev[\phi^{(t)}_k] \leq \alpha$, for each $k\in [K]$. 
\end{remark}
\begin{remark} 
The use of randomization (strictly) enhances the statistical efficiency of the prediction sets while preserving the theoretical validity of the procedure. Importantly, if one opts to avoid randomization, for instance, to mitigate any potential for p-hacking, then setting $u^{(t)}=0$ maintains the coverage guarantee, which is ensured by  Markov’s inequality.
\end{remark}
\begin{remark}\label{rem:ass}
    On examining the proof, it is easy to see that the degradation of the coverage degrades smoothly with the degree of violation of Assumption~\ref{ass:1}. In particular, if we relax the assumption in \eqref{eq:ass-neg-tot-corr} to $\sum_{k=1}^K \Ev[\phi_k^{(t)}W_k^{(t)}] \leq (1+\nu) \sum_{k=1}^K \Ev[\phi_k^{(t)}]\Ev[W_k^{(t)}]$, for some $\nu \geq 0$ which measures the degree to which assumption \eqref{eq:ass-neg-tot-corr} is violated, we see that the corresponding bound on the miscoverage becomes $2\alpha(1+\nu)$, recovering the above theorem when $\nu=0$. 
\end{remark}

     A simple instance in which the negative total elementwise correlation stated in \eqref{eq:ass-neg-tot-corr} holds is described here. Consider the scenario involving only two models, resulting in a weight vector denoted by $(W_1, 1-W_1)$, where the superscript $(t)$ is omitted for simplicity. Moreover, assume $\phi_1=\phi_2=\phi$, which might occur when two algorithms deliver nearly identical results.
     If $\mathrm{cov}(W_1, \phi) = \rho$, then it follows that $\mathrm{cov}(1-W_1, \phi) = -\rho$; and this implies $\sum_{k=1}^2\Ev[\phi W_k] = \sum_{k=1}^2 \Ev[\phi]\, \Ev[W_k]$, so \eqref{eq:ass-neg-tot-corr} holds with equality. This is just a simple illustration; the assumption is investigated later.

If the weight vector takes the form $w^{(t)}_k \approx 1$, and $w^{(t)}_j \approx 0$ for all $j \neq k$, then the procedure effectively reduces to selecting the $k$-th model, which is the desired result when there is a clearly best model. A potential drawback of the procedure is that it may produce a union of intervals, even when the input sets are intervals. Luckily, the combined set is an interval as long as the intersection of the $K$ intervals is non-empty \citep{gasparin2024}, which we expect to be the case more often than not. In fact, the probability of observing either an empty set or a union of sets is close to zero in our experiments. Moreover, when the weights are concentrated on a single algorithm, the majority vote reduces to the set of that algorithm.

As a final remark, we note that the expectation in the marginal coverage condition in \eqref{eq:ass-neg-tot-corr} is taken over both the training data and the test point. An alternative condition is the following:
\begin{equation} \label{eq:ass-cond-cov}
        \Ev\left[\phi_{k}^{(t)} \mid \{ Z^{(i)}\}_{i=1}^{t-1}\right] \leq \alpha , \text{ for all } k \in [K],
        \end{equation}
where, in this case, the miscoverage error is bounded conditionally on the ``training set'' $\{Z^{(i)}\}_{i=1}^{t-1}$. In practice, the model may not have been updated after each additional point, but we still call this condition as \emph{training conditional coverage}. This condition can be used to extend the results in Theorem~\ref{prop:cov}.
\begin{lemma}\label{lemma:cond-cov}
    Let $\setc_1^{(t)}, \dots, \setc_K^{(t)}$ be $K \geq 2$ different prediction sets for $Y^{(t)}$, let $W^{(t)}$ be a random vector in $\Delta_{K-1}$ depending only on $\{Z^{(i)}\}_{i=1}^{t-1}$, and consider the weighted majority vote set $\setc_M^{(t)}$ in~\eqref{eq:cm}. If \eqref{eq:ass-cond-cov} holds, then $\pr(Y^{(t)} \notin \setc_M^{(t)}) \leq 2\alpha$.
\end{lemma}
\begin{proof}
    Following the same steps as in Theorem~\ref{prop:cov}, if the assumption holds, then
    \[
    \begin{split}
    \pr\left(Y^{(t)} \notin \setc_M^{(t)}\right) &\leq 2 \sum_{k=1}^K \Ev \left[ W_k^{(t)} \phi_k^{(t)} \right] = 
     2 \sum_{k=1}^K \Ev\left[ W_k^{(t)} \Ev\left[\phi_k^{(t)} \mid \{Z^{(i)}\}_{i=1}^{t-1} \right] \right] \leq 2\alpha.
    \end{split}
    \]
\end{proof}
However, this condition does not hold for conformal prediction methods.
A weaker type of guarantee is addressed by PAC-type results. In particular, it involves two parameters: the miscoverage rate $\alpha$ and the parameter $\delta$. The training conditional coverage for the split conformal method is explored in \cite{vovk2012}, whereas the training conditional coverage for the jackknife+ and the $K$-fold CV+ \citep{barber2021} is examined in \cite{bian2023}. In particular, under certain conditions, the training conditional coverage of these methods is  asymptotically (in the size of the calibration set) $1-\alpha$ with probability at least $1-\delta$. 

A different condition holds for full conformal prediction in the online case when data is iid (see Chap. 2.2 in \cite{vovk2005}), specifically
\begin{equation}
\label{eq:cond2}
    \Ev\left[\phi_k^{(t)} \mid \{\phi_k^{(i)}\}_{i=1}^{t-1}\right] \leq \alpha, \mathrm{~for~all~}k \in [K].
\end{equation}
The left hand side above conditions on less information than~\eqref{eq:ass-cond-cov}.
This condition states that the errors are independent in the online setting when data are independent and identically distributed. However, this condition does not suffice to guarantee a miscoverage rate of at most $2\alpha$.

\section{The COMA meta-algorithm}
\label{sec:dynamic}
So far, we have studied the theoretical properties of the method in the case where the weights themselves are random variables; what remains is to find a sensible and practical way to update these weights in order to improve the statistical efficiency of our weighted majority vote set. To solve this, we will employ a particular variant of the exponential weighted majority (\textsc{ew}) algorithm. To that end, we must first define the loss functions we will use within that algorithm. Often (or even typically) \textsc{ew} is applied to bounded losses, but our losses below can be unbounded, but their nonnegativity will allow us to adapt the AdaHedge algorithm by~\cite{derooij2014} for our purposes. We expand below.

\subsection{Loss function definition}
We begin by defining the loss functions relevant to the problem under consideration: for regression tasks, the loss is quantified by the interval's size, and for classification tasks, by the count of elements within the set. In particular, we define the loss function $\ell(\cdot)$ as 
\begin{equation}\label{eq:loss_fn}
    \ell := \ell(\setc) = 
    \begin{cases}
       \text{Lebesgue measure}(\setc), \quad &\mathrm{if}~\mathcal{Y} \subseteq \mathbb{R},\\
       |\setc|, \quad &\mathrm{if}~\mathcal{Y} = \{y_1, \dots, y_D\},
    \end{cases}
\end{equation}
where $\setc$ denotes any set, $|\setc|$ represents the cardinality of $\setc$, and $\{y_1, \dots, y_D\}$ describes a discrete set with $D$ elements (or labels). 
 {In general, one could choose some nondecreasing function of the Lebesgue measure (or of the cardinality), but the Lebesgue measure and the cardinality seem to be a sensible canonical choice, and we therefore adopt it throughout this paper. }

At each iteration, we observe losses of experts $\ell^{(t)} = (\ell_1^{(t)}, \dots, \ell_K^{(t)}) \in\{\real^+\}^K$ and the cumulative loss for a given model after $t$ rounds is given by $L_k^{(t)} := \ell^{(1)}_k+ \dots + \ell^{(t)}_k$. A notable property of the majority vote is that it generates sets that are never larger than twice the weighted average size of the initial sets. 
\begin{lemma}
    \label{lemma:length}
    Let $\setc_M^{(t)}$ be the set defined in \eqref{eq:cm}, $\ell_M^{(t)} = \ell(\setc_M^{(t)})$ be the loss function defined in \eqref{eq:loss_fn} associated with $\setc_M^{(t)}$, and $w^{(t)}$ be any realization of $W^{(t)}$. Then,
    \begin{align}
    \label{eq:length_cm}
        \ell_M^{(t)} = \ell(\setc_M^{(t)}) &\leq 2\sum_{k=1}^K w_k^{(t)}\ell(\setc_k^{(t)}), \text{ and  }\\
    \label{eq:avg_l}
        \Ev\left[\ell_M^{(t)}\right] &\le 2 \ln 2 \,\Ev\left[\sum_{k=1}^K W_k^{(t)}\ell_k^{(t)}\right].
    \end{align}
\end{lemma}
\begin{proof}
    We start with the discrete case $\mathcal{Y} = \{y_1, \dots, y_D\}$. In particular, let $\setc_1^{(t)}, \dots, \setc_K^{(t)}$ be $K$ different sets and let $w^{(t)}$ be any realization of $W^{(t)}$; then, we have that
    \[
    \begin{split}
        \ell_M^{(t)} &= \sum_{d=1}^D \ind\left\{\sum_{k=1}^K w_k^{(t)} \ind\left\{y_d \in \setc_k^{(t)}\right\} > \frac{1}{2} + \frac{u^{(t)}}{2}\right\} \leq \sum_{d=1}^D \ind\left\{\sum_{k=1}^K w_k^{(t)} \ind\left\{y_d \in \setc_k^{(t)}\right\} > \frac{1}{2} \right\}\\
        &\stackrel{(i)}{\leq} \sum_{d=1}^D 2 \sum_{k=1}^K w_k^{(t)} \ind\left\{y_d \in \setc_k^{(t)} \right\}
        = 2 \sum_{k=1}^K w_k^{(t)} \sum_{d=1}^D \ind\left\{y_d \in \setc_k^{(t)} \right\} = 2 \sum_{k=1}^K w_k^{(t)} \ell(\setc_k^{(t)}),
    \end{split} 
    \]
    where $(i)$ holds due to the fact that $\ind\{x > 1\} \leq x$, if $x\geq 0$. The proof in the case $\mathcal{Y} \subseteq \mathbb{R}$ is identical; it replaces the summation over $d$ by an integral. 
    Similarly, we have 
    \[
    \begin{split}
        \Ev\left[\ell_M^{(t)}\right] &= \Ev\left[\sum_{d=1}^D \ind\left\{\sum_{k=1}^K W_k^{(t)} \ind\left\{y_d \in \setc_k^{(t)}\right\} > \frac{1}{2} + \frac{U^{(t)}}{2}\right\}\right]\\
        &=\Ev\left[\Ev\left[\sum_{d=1}^D \ind\left\{\sum_{k=1}^K W_k^{(t)} \ind\left\{y_d \in \setc_k^{(t)}\right\} > \frac{1}{2} + \frac{U^{(t)}}{2}\right\}\mid U^{(t)}\right]\right]\\
        &\le \Ev\left[\sum_{d=1}^D\Ev\left[\frac{2}{1+U^{(t)}} \sum_{k=1}^K W_k^{(t)} \ind\left\{y_d \in \setc_k^{(t)}\right\}\mid U^{(t)}\right]\right]\\
        &= \Ev\left[\frac{2}{1+U^{(t)}} \Ev\left[\sum_{k=1}^K W_k^{(t)}\ell_k^{(t)}\right]\right] = 2 \ln 2 \,\Ev\left[\sum_{k=1}^K W_k^{(t)}\ell_k^{(t)}\right],
    \end{split} 
    \]
 {The proof in the case $\mathcal{Y} \subseteq \mathbb{R}$ is analogous, due to the Fubini-Tonelli Theorem.\\(Note that $\ln2 \approx 0.7 < 1$)}    
\end{proof}

 {From \eqref{eq:length_cm} and \eqref{eq:avg_l}, it is now clear that our goal is to find a good weighting system in order to narrow down the size of $\setc_M^{(t)}$, in particular, we aim to assign a higher weight to experts that consistently yield smaller intervals compared to others. In other words, we want to prioritize prediction algorithms that yield smaller sets. Furthermore, we want the cumulative loss of our proposed method $L_M^{(t)} := \ell_M^{(1)} + \dots + \ell_M^{(t)}$ to be not much larger than (and possibly smaller than) the cumulative loss incurred by the best expert in hindsight $L_*^{(t)} := \min_{k \in [K]} L_k^{(t)}$. As a remark, the factor $\ln 2$ in \eqref{eq:avg_l}, can be seen as the improvement due to randomization.}

\subsection{Employing AdaHedge within COMA}
Our proposed way to update weights is to use the \textsc{ew} (or Hedge) Algorithm~\citep{littlestone1994,freund1997}, which corresponds to Algorithm~\ref{alg:ewa} with $\eta^{(t)}=\eta$, for all $t$. A key quantity is the \emph{hedge} loss $H^{(t)} = h^{(1)} + \dots + h^{(t)}$, where $h^{(t)}$ is the dot product $h^{(t)} = w^{(t)} \cdot \ell^{(t)} = \sum_k w_k^{(t)} \ell_k^{(t)}$. This can be interpreted as the expected loss achieved by randomly selecting the model $k$ with probability $w_k^{(t)}$.

\begin{algorithm}
\caption{Conformal Online Model Aggregation}
\label{alg:ewa}
\begin{algorithmic}[1]
\Require{$\setc_1^{(t)}, \dots, \setc_K^{(t)}$ at each round $t$, initial learning rate $\eta^{(0)} \geq 0$}
\State {$w_k^{(1)} \leftarrow 1/K, L_k^{(0)} \leftarrow 0 ,$ $k=[K]$}\;
\For{rounds $t=1,\dots,T$}
    \State $\setc_M^{(t)} \leftarrow \left\{y \in \mathcal{Y}: \sum_{k=1}^K w_k^{(t)} \ind\{y \in \setc_k^{(t)} \} > \frac{1+u^{(t)}}{2}\right\}$\;
    \State Receive loss $\ell_k^{(t)}$, update $L_k^{(t)} := L_k^{(t-1)} + \ell_k^{(t)}$\;
    \State Update learning rate $\eta^{(t)}$\;
    \State $w_k^{(t+1)} \leftarrow 
    \exp\{-\eta^{(t)} L_k^{(t)}\}/\sum_{j=1}^K\exp\{-\eta^{(t)} L_j^{(t)}\}$\;
\EndFor
\end{algorithmic}
\end{algorithm}

The algorithm adjusts the weights based on the performance of the methods in various rounds, with adjustments made in inverse proportion to the exponential of their total loss scaled by a parameter $\eta$. This ``learning rate'' $\eta$ plays a crucial role; if $\eta$ approaches zero, the weights approach a uniform weighting, and if $\eta \to \infty$ the algorithm reduces to the Follow-the-Leader strategy, which puts all the weight on the method with the smallest loss so far.

In certain scenarios, determining the appropriate value for this parameter may be difficult; indeed, we do not know if there is a method outperforming the others or any constant works well in some situations but not in others. 
Our solution is to use the AdaHedge algorithm, introduced in \cite{derooij2014}, which dynamically adjusts the learning parameter over time. At each round, weights are assigned according to $w^{(t+1)}_k \propto e^{-\eta^{(t)} L^{(t)}_k}, k \in [K]$, with 
\begin{equation*}
    \eta^{(t)} := \frac{\ln K}{\delta^{(1)} + \dots + \delta^{(t-1)}},
\end{equation*} 
where $\delta^{(t)} = h^{(t)} - m^{(t)}$ 
is the difference between the hedge loss $h^{(t)}$ and the \emph{mix} loss, 
\begin{equation*}
    m^{(t)} := -\frac{1}{\eta^{(t)}} \ln\left(w^{(t)} \cdot \exp\{-\eta^{(t)} \ell^{(t)}\}\right).
\end{equation*}

The AdaHedge algorithm provides an upper bound on regret, defined as the difference between the hedge loss and the loss of the best method up to time $t$, specifically: $H^{(t)} - L_*^{(t)}$.

Consider $\ell_+^{(t)}=\max_{k \in [K]} \ell_k^{(t)}$ and $\ell_-^{(t)}=\min_{k\in[K]} \ell_k^{(t)}$ to represent the smallest and largest loss in round $t$, and let $L_+^{(t)}$ and $L_-^{(t)}$ represent their cumulative sum. Additionally, define $$V^{(t)}:=\max\{v^{(1)}, \dots, v^{(t)}\}$$ denote the largest loss range, where $v^{(t)} = \ell_+^{(t)} - \ell_-^{(t)}$. 
We now bound the loss of the sets produced by the majority vote method. 

\begin{theorem}\label{th:bound}
If Assumption~\ref{ass:1} holds, then the COMA algorithm satisfies $\pr(Y^{(t)} \in \setc_{M}^{(t)}) \ge 1-2\alpha.$
The loss of the weighted majority vote set over $T$ rounds can be bounded as
\begin{equation}\label{eq:bound-cl}
L_M^{(T)} \leq  2L_*^{(T)} + 4 \left(V^{(T)} \ln K \frac{(L_+^{(T)} - L_*^{(T)})(L_{*}^{(T)} - L_-^{(T)})}{L_+^{(T)} - L_-^{(T)}} \right)^{1/2} + 2 V^{(T)}\left(\frac{16}{3} \ln K + 2 \right).
\end{equation}
In addition, if the loss range is bounded, i.e. $\sup_t v^{(t)} \le D$, then 
\begin{equation}\label{eq:regr_as}
    \frac{L_M^{(T)}}{T} \leq  \frac{2L_*^{(T)}}{T}+o(1)
\end{equation}
\end{theorem}
\begin{proof}
    The first part of the theorem is just a consequence of Theorem~\ref{prop:cov} applied to Algorithm~\ref{alg:ewa}.
    From Lemma~\ref{lemma:length} we have $\ell_M^{(t)} \leq 2 h^{(t)}$. This implies that $L_M^{(T)} = \ell_M^{(1)} + \dots + \ell_M^{(T)} \leq 2 H^{(T)}$.
    Next, by Theorem 8 in \citet{derooij2014}, we have
\begin{equation}\label{eq:regret-bound}
H^{(T)}   \leq L_*^{(T)} +  2 \left(V^{(T)} \ln K \frac{(L_+^{(T)} - L_*^{(T)})(L_{*}^{(T)} - L_-^{(T)})}{L_+^{(T)} - L_-^{(T)}} \right)^{1/2} + V^{(T)}\left(\frac{16}{3} \ln K + 2 \right).
\end{equation}
    Combining the earlier two statements  yields the desired claim. If the loss range is bounded, then $v^{(t)}\le 
    D$, and using Corollary 9 in \citet{derooij2014} we have 
    \[
    \begin{split}
    L_M^{(T)} & \leq 2L_*^{(T)} +  4\left(\sum_{t=1}^T(v^{(t)})^2 \ln K \right)^{1/2} + 2V^{(T)}\left(\frac{4}{3} \ln K + 2 \right)\\
    &\le 2L_{*}^{(T)} + 4(TD^2\ln K)^{1/2} + 2D\left(\frac{4}{3} \ln K + 2 \right)
    \end{split}
    \]
    Dividing both sides by $T$ gives the result.
\end{proof}
The result in \eqref{eq:bound-cl} is identical to the regret bound~\eqref{eq:regret-bound}, except for a multiplicative factor of 2.  {In addition, under the assumption of bounded loss range (which is automatically satisfied in the classification setting, for example), the empirical average loss of the majority vote set is at most twice the empirical average loss of the best method in hindsight, up to an asymptotically vanishing term. Although initially developed for aggregating point forecasts, we adapted the AdaHedge algorithm to a distinctly different context: the aggregation of sets. It is somewhat surprising that the regret bound can also be applied to this scenario.  Clearly, other methods are available to update the weights. However, we recommend the proposed solution because it does not require any tuning parameters (it is completely data-driven), ensures a regret bound even in the case of unbounded loss functions, and demonstrates strong empirical performance. When our dynamic merging algorithm is used (with constant, adaptive or another stepsize rule) in this conformal context, we call the method as ``conformal online model aggregation'', or COMA for short.}

As outlined in Theorem~\ref{th:bound}, Assumption~\ref{ass:1} guarantees a miscoverage rate of at most $2\alpha$ for the proposed solution. However, since the weights are updated based on the losses of the different algorithms, a sufficient condition which lies between \eqref{eq:ass-cond-cov} and \eqref{eq:cond2} and guarantees a miscoverage rate of at most $2\alpha$ for Algorithm~\ref{alg:ewa} is
\[
\Ev\left[\phi_k^{(t)} \mid \{\phi_k^{(i)}\}_{i=1}^{t-1}, \{\ell_k^{(i)}\}_{i=1,k=1}^{t-1,K}\right] \leq \alpha, \mathrm{~for~all~}k \in [K].
\]
 {However, it remains unclear whether this condition is likely to hold either in theory or in practice. For this reason, we regard Assumption~\ref{ass:1} as the most reasonable choice in practical applications. From a theoretical point of view, this condition is difficult to verify, since the sets are based on the same underlying data and both the errors and the losses may be arbitrarily correlated. Nevertheless, it can be empirically assessed and appears to hold reasonably well in the experiments presented in Section~\ref{sec:dyn_real}. In addition, if the assumption is not met, Remark~\ref{rem:ass} shows that the coverage guarantee weakens gradually as the violation grows.}

In certain instances, the Lebesgue measure of the set cannot be directly employed because of situations where the interval corresponds to entire real line. For example, we will see that in adaptive conformal prediction \citep{gibbs2021}, it may happen that the predictive set of an agent coincides with $\mathcal{Y}$. A possible solution is to define the loss function as an increasing, nonnegative, bounded function of the measure of the set (so that the weight of a predictor does not get permanently set to zero if its loss happens to be unbounded in some round).
We will discuss this example in Section~\ref{subsec:aci-2}.

\section{COMA under distribution shift}
\label{sec:dyn_aci}
 {Up to this point, we have focused on the iid setting and described the properties of the COMA in this scenario. As we will see in Section~\ref{sec:dyn_real}, which reports the experimental results, under the iid setting, COMA tends to concentrate the weights on the best-performing model. However, what occurs if distributional shifts arise and different models exhibit varying levels of performance at different points in time?} First of all, as discussed in Section~\ref{sec:problem_setup}, conformal prediction provides a general framework to transform the output of any predictive algorithm into a prediction set; however, its validity is based on the assumption of independence and identical distribution (or exchangeability). This assumption is violated when the underlying data distribution shifts over time and standard conformal inference can no longer be expected to yield reliable guarantees. To address this challenge, several methods have been proposed in recent years, including \citet{gibbs2021} and \citet{angelopoulos2023pid}.  {We now briefly outline their approaches before extending our COMA procedure to this setting. In particular, we will see that COMA can be used in combination with these approaches, serving as a flexible tool in the considered scenario. For clarity of exposition, we first introduce the methods in the case of a single algorithm, deferring the multi-expert extensions to Section~\ref{subsec:wrapper-aci} and Section~\ref{subsec:aci-2}.}

\subsection{Adaptive conformal inference methods}
In \citet{gibbs2021}, \emph{adaptive conformal inference} (ACI) is introduced as a novel method for constructing prediction sets in an online manner, providing robustness to changes in the marginal distribution of the data. 
Their work is based on the update of the error level $\alpha$ to achieve the desired level of confidence. Indeed, given the possible non-stationary nature of the data-generating distribution, conventional results do not guarantee $1-\alpha$ coverage. However, at each time $t$, an alternative value $\alpha^{(t)}$ might exist, allowing the attainment of the desired coverage. 


The procedure described in \citet{gibbs2021} recursively updates the error rate $\alpha$ as follows:
\begin{gather}
\label{eq:alpha_st}
    \alpha^{(1)} = \alpha,\\
\label{eq:alpha_up}
    \alpha^{(t)} = \alpha^{(t-1)} + \gamma (\alpha - \phi^{(t-1)}), \quad t\geq2,
\end{gather}
where $\phi^{(t)} = \ind\{y^{(t)} \notin \setc^{(t)}(\alpha^{(t)})\}$ is the sequence of miscoverage events setting the miscoverage rate to $\alpha^{(t)}$, while $\gamma>0$ represents a step size parameter. We refer to Appendix~\ref{sec:app_ci} for a more detailed explanation of the method. Possible improvements are proposed in~\citet{gibbs2023} and~\citet{zaffran2022}; in particular, their proposals tune the parameter $\gamma$ over time using an online learning procedure. A drawback arises from the fact that in certain situations, ACI can return either $\mathcal{Y}$ or $\emptyset$ as a set, attributed to the circumstance that in some iterations, the parameter $\alpha^{(t)}$ can be smaller than 0 or greater than 1.

Another method is presented in \citet{angelopoulos2023pid}, where instead of updating the error level $\alpha$, there is an update of the quantile of the distribution of the \emph{scores}. In this setting, we define $s^{(t)}: \mathcal{X} \times \mathcal{Y} \to \mathbb{R}$ as the \emph{conformal score}, which measures the accuracy of the prediction at time $t$. If different models are involved one can use different scores (e.g. residual or density based scores). The procedure, starting from an initial threshold $q^{(1)} = q$, updates the quantiles as follows:
\begin{gather*}
    q^{(t)} = q^{(t-1)} + \gamma(\phi^{(t-1)} - \alpha), \quad t \geq 2,
\end{gather*}
where, in this case, $\phi^{(t)}=\ind\{y^{(t)} \notin \setc^{(t)}(q^{(t)})\}$ and the sets at each time are simply defined by
\begin{equation}\label{eq:qt_set}
    \setc^{(t)}(q^{(t)}) = \left\{y \in \mathcal{Y}: \,s^{(t)}(x^{(t)}, y) \leq q^{(t)}\right\}.
\end{equation}
As noted in \citet{angelopoulos2023pid}, both ACI and quantile tracking, can be rephrased to an online descent algorithm with respect to the pinball loss \citep{koenker2005}. In addition, both methods achieve a \emph{long-run coverage} in time, defined as
\begin{equation}\label{eq:long-run}
    \frac{1}{T} \sum_{t=1}^T \phi^{(t)} = \alpha + o(1),
\end{equation}
under few or no assumptions on $\{z^{(t)}\}_{t=1}^T$ and $\{s^{(t)}\}_{t=1}^T$, where $o(1)$ is a quantity that tends to zero as $T \to \infty$. In particular, \eqref{eq:long-run} holds deterministically and is different from the probabilistic assumption in~\eqref{eq:coverage} valid in an iid\ setting. Improvements are also suggested for the quantile tracking method, such as an adaptive learning rate or the introduction of an \emph{error integrator}. The proposal in~\citet{angelopoulos2024decaying} is to set a decaying step size $\gamma^{(t)}$. They show that if scores are bounded and $\gamma^{(t)} \propto t^{-1/2-\epsilon}$ for some $\epsilon \in (0,1/2)$, then the long-run coverage is bounded as $O(\frac{1}{T^{1/2-\epsilon}})$ in the adversarial setting, while in the iid case the algorithm achieves a coverage convergence guarantee. 

In light of the dynamics shift observed in the marginal distribution of the data, it is conceivable that the weights assigned to different algorithms may also undergo variations throughout iterations, since different models may be preferable at different time points. This feature makes our method particularly appealing. We now describe two different ways to combine our dynamic ensembling algorithm with ACI: either as a wrapper that does not alter the inner workings of ACI (Section~\ref{subsec:wrapper-aci}), or by altering the feedback provided to ACI itself (Section~\ref{subsec:aci-2}). 

\subsection{Decentralized COMA under distribution shift}
\label{subsec:wrapper-aci}
In the scenario described above, suppose that each agent operates independently by providing an interval at each iteration and, subsequently, the \emph{aggregator} merges these intervals according to the observed vector $w^{(t)}$. This framework can be referred to as a decentralized dynamic merging. In this setting, the advantage of the quantile tracking method is that it never returns sets of infinite measure. This implies that the loss function used can remain the size of the different sets, ensuring that the bound described in~\eqref{eq:bound-cl} is valid. In the case of using the described ACI approach, it becomes necessary to employ a function $g(\cdot)$ that is capable of returning a finite number when $\alpha_k^{(t)} \leq 0$.
Before proceeding with the rest of the section, we introduce a proposition which holds under the following assumptions:
\begin{enumerate}
    \item Negative total elementwise empirical correlation between $\{\phi^{(t)}\}_{t=1}^T$ and $\{w^{(t)}\}_{t=1}^T$:
    \begin{equation}\label{eq:ass1}
        \frac1T \sum_{t=1}^T \sum_{k=1}^K (\phi_k^{(t)} - \Bar{\phi}_k)(w_k^{(t)} - \Bar{w}_k) \leq 0, 
    \end{equation}
    where $\Bar{\phi}_k:= \frac1T \sum_{t=1}^T \phi_{k}^{(t)}$ and $\Bar{w}_k:= \frac1T\sum_{t=1}^T w_{k}^{(t)}$.
    \item No randomization: the value $u^{(t)}$ in \eqref{eq:cm} equals 0:
    \begin{equation}\label{eq:ass2}
        u^{(t)} = 0, \quad t=1, \dots, T.
    \end{equation}
\end{enumerate}

\begin{proposition}\label{prop:long_run}
    Let $\{z^{(t)}=(x^{(t)}, y^{(t)})\}_{t=1}^T$ be an arbitrary sequence of data points, $\{w^{(t)}=(w_1^{(t)}, \dots, w_k^{(t)})\}_{t=1}^T$ be a sequence of weight vectors, and $\{\phi_k^{(t)}\}_{t=1}^T$ be the sequence of miscoverage events for the $k$-th algorithm, $k\in[K]$. Under assumptions \eqref{eq:ass1} and \eqref{eq:ass2}, for all $\alpha \in (0, 1/2]$, decentralized COMA in conjunction with adaptive conformal inference \eqref{eq:alpha_up} satisfies
    \begin{equation}\label{eq:long_run}
    \frac{1}{T} \sum_{t=1}^T \ind\{y^{(t)} \notin \setc_M^{(t)}\} \leq 2\alpha + 2 \frac{1-\alpha+\gamma}{T\gamma}.
    \end{equation}
\end{proposition}
    
\begin{proof}
    We remark that, if $u^{(t)}=0$, then $y^{(t)}$ is not contained in $\setc^{(t)}$ if and only if $\sum_{k=1}^K w_k^{(t)} \ind\{y^{(t)} \notin \setc_k^{(t)}(\alpha_k^{(t)})\} \geq 1/2$. Then, since $\phi_k^{(t)}=\ind\{y^{(t)} \notin \setc_k^{(t)}(\alpha_k^{(t)})\}$, we can write
    \[
    \begin{split}
        \sum_{t=1}^T \ind\{y^{(t)} \notin \setc_M^{(t)}\} &= \sum_{t=1}^T \ind\left\{\sum_{k=1}^K w_k^{(t)}\ind\{y^{(t)} \notin \setc_k^{(t)}(\alpha_k^{(t)})\} \geq \frac{1}{2} \right\}\\
        & \stackrel{(i)}{\leq} \sum_{t=1}^T 2 \sum_{k=1}^K w_k^{(t)} \ind\{y^{(t)} \notin \setc_k^{(t)}(\alpha_k^{(t)}) \},\\
        &= 2 \sum_{k=1}^K \sum_{t=1}^T w_k^{(t)} \phi_k^{(t)}\\
        & \stackrel{(ii)}{\leq} 2 \sum_{k=1}^K \frac{1}{T} \left(\sum_{t=1}^T w_k^{(t)} \right) \left(\sum_{t=1}^T \phi_k^{(t)} \right) \\
        & \stackrel{(iii)}{\leq} 2 \sum_{k=1}^K \frac{1}{T} \left( \sum_{t=1}^T w_k^{(t)} \right) \left(T\alpha + \frac{1-\alpha+\gamma}{\gamma} \right)\\
        &= 2\left(\alpha + \frac{1-\alpha+\gamma}{T\gamma} \right) \sum_{k=1}^K \sum_{t=1}^T w_k^{(t)}\\
        &= 2T\alpha + 2\frac{1-\alpha+\gamma}{\gamma},\\
    \end{split}
    \]
    where $(i)$ holds due to the fact that $\ind\{x \geq 1\} \leq x,$ if $x\geq 0$ and $(ii)$ is due to assumption \eqref{eq:ass1}, in fact if the assumption holds, then $T \sum_{k=1}^K \sum_{t=1}^T w_k^{(t)} \phi_k^{(t)} - \sum_{k=1}^K(\sum_{t=1}^T \phi_k^{(t)}) (\sum_{t=1}^T w_k^{(t)}) \leq 0$. Proposition 4.1 in \citet{gibbs2021} along with $\alpha \in (0,1/2]$ implies $(iii)$. Dividing both sides by $T$ gives the result.
\end{proof}
Similar results can be obtained using the quantile tracking method under the assumption of bounded scores (see Proposition 1 in \citet{angelopoulos2023pid} and Theorem 1 in \citet{angelopoulos2024decaying}). Randomization is not included because the above is a deterministic result; however, the empirical results using randomization appear to be reasonably good also in this setting. The result \eqref{eq:long_run} guarantees long-run coverage at least of level $1-2\alpha$ for any distribution, provided specific conditions about the observed weights and miscoverage errors are met. In our case the weights will be obtained using the method described in Section~\ref{sec:dynamic}.  {Although the assumption in \eqref{eq:ass1} is difficult to establish rigorously, we will see that it is generally satisfied in practice.}

 {The algorithm is presented in Algorithm~\ref{alg:ewa_q}, and it requires an additional step to update the quantiles after observing the true response value at each iteration. This update can be performed independently by each expert in a decentralized setting, where the experts work separately and the aggregator only receives the $K$ sets at each iteration. The algorithm can be modified using the miscoverage rate $\alpha$ instead of the quantiles.}

\begin{algorithm}
\caption{Decentralized COMA under distribution shift}
\label{alg:ewa_q}
\begin{algorithmic}[1]
\Require Initial learning rate $\eta^{(0)} \geq 0$ and starting quantiles $q_1,\dots,q_K$
\State $w_k^{(1)} \leftarrow 1/K, L_k^{(0)} \leftarrow 0 ,$ $k=[K]$
\State $q_k^{(1)}=q_k,\, k=[K]$
\For{rounds $t=1,\dots,T$}
    \State Require $\setc_k^{(t)}(q_k^{(t)}), \, k \in [K]$ 
    \State $\setc_M^{(t)} \leftarrow \left\{y \in \mathcal{Y}: \sum_{k=1}^K w_k^{(t)} \ind\{y \in \setc_k^{(t)} \} > \frac{1}{2}\right\}$
    \State Receive loss $\ell_k^{(t)}$, update $L_k^{(t)} := L_k^{(t-1)} + \ell_k^{(t)}$
    \State Update learning rate $\eta^{(t)}$
    \State $w_k^{(t+1)} \leftarrow 
    \exp\{-\eta^{(t)} L_k^{(t)}\}/\sum_{j=1}^K\exp\{-\eta^{(t)} L_j^{(t)}\}$
    \State Observe $y^{(t)}$ and miscoverage errors for every experts $\phi_k^{(t)}, k\in [K]$
    \State Update quantiles $q_k^{(t+1)}, k \in [K]$
\EndFor
\end{algorithmic}
\end{algorithm}

\subsection{Adaptive conformal inference directly applied on dynamic merging}\label{subsec:aci-2}

The approach described in the preceding section had agents updating the quantiles of their score function based on \emph{their own} past errors. This makes sense when all agents can observe the ground truth and calculate their own errors, and they do not care about the aggregator's goals. In other words, the dynamic merging algorithm was just an outer wrapper that did not interfere with the inner functioning of the $K$ adaptive conformal inference algorithms.

Below, we show that our dynamic ensembling method can provide direct feedback to the adaptive conformal inference framework, and this can have some advantages in settings where the end goal is good aggregator performance only, and individual agents do no have their own goals (of maintaining coverage, say). First, we observe that if the absolute residuals are employed as the score function and if the quantile ($q^{(t)})$ is common across all algorithms, then the size of the sets at time $t$ is equal for all intervals. This implies that the weights are not updated during the various iterations. In this case, we propose using a value $\alpha^{(t)}$ that is common for all $K$ agents, and it is updated by the aggregator based on the performance of the set obtained by the exponential weighted majority vote. In other words, the aggregator, at each iteration, tries to learn the coverage level required to obtain a $(1-\alpha)$-prediction set. The procedure updates the $\alpha$-level according to the miscoverage event $\phi^{(t)}= \ind\{y^{(t)} \notin \setc_M^{(t)}(\alpha^{(t)})\}$, where 
\[
\setc_M^{(t)}(\alpha^{(t)}) := \left\{y \in \mathcal{Y}: \sum_{k=1}^K w_k^{(t)} \ind\left\{y \in \setc_k^{(t)}\left(\alpha^{(t)}\right)\right\} > 1/2 \right\},
\]
and $w_k^{(t)}$ are the weights learned by the COMA procedure.
The error level (common among agents) is updated as in \eqref{eq:alpha_st} and \eqref{eq:alpha_up}. It is important to note that the interval lengths can be different between the various agents even if $\alpha^{(t)}$ is common.
 {In such a scenario, ACI operates directly on the set produced by the aggregator, and so we directly recover the properties of the ACI method. This is due to the fact that the sets are nested with respect to the parameter $\alpha^{(t)}$ and the set is $\mathcal{Y}$ if $\alpha^{(t)}\le 0$ and $\emptyset$ if $\alpha^{(t)}\ge 1$ (see \cite{szabadvary2026beyond}). Further, there is no need for randomization to enhance the length of the intervals. This scenario is plausible when the different agents (models) can collaborate with each other, and the aggregator is able to update both the weights and the miscoverage rate.}

 {The procedure is outlined in Algorithm~\ref{alg:ewa_a}, where the $\alpha$-level is shared among the $K$ models and updated by the aggregator itself, based on its own miscoverage events.}

\begin{algorithm}
\caption{Centralized COMA under distribution shift}
\label{alg:ewa_a}
\begin{algorithmic}[1]
\Require{Initial learning rate $\eta^{(0)} \geq 0$ and miscoverage rate $\alpha$}
\State {$w_k^{(1)} \leftarrow 1/K, L_k^{(0)} \leftarrow 0 ,$ $k=[K]$}\;
\State {$\alpha^{(1)}=\alpha$}
\For{rounds $t=1,\dots,T$}
    \State Require $\setc_k^{(t)}(\alpha^{(t)}), \, k \in [K]$;
    \State $\setc_M^{(t)} \leftarrow \left\{y \in \mathcal{Y}: \sum_{k=1}^K w_k^{(t)} \ind\{y \in \setc_k^{(t)} \} > \frac{1}{2}\right\}$\;
    \State Receive loss $\ell_k^{(t)}$, update $L_k^{(t)} := L_k^{(t-1)} + \ell_k^{(t)}$\;
    \State Update learning rate $\eta^{(t)}$\;
    \State $w_k^{(t+1)} \leftarrow 
    \exp\{-\eta^{(t)} L_k^{(t)}\}/\sum_{j=1}^K\exp\{-\eta^{(t)} L_j^{(t)}\}$\;
    \State Observe $y^{(t)}$ and miscoverage event $\phi^{(t)} = \ind\{y^{(t)} \notin \setc_M^{(t)}\}$
    \State Update miscoverage rate $\alpha^{(t+1)}$\;
\EndFor
\end{algorithmic}
\end{algorithm}

\section{Experimental results}
\label{sec:dyn_real}
We now test the approaches proposed in the previous sections in different scenarios. In the iid\ setting, it is often the case that our adaptive stepsize update rule results in the weight vector rapidly putting almost full (unit) weight on the best predictor, resulting effectively in online model ``selection''. In contrast, in non-iid settings with possible distribution shifts, we will see that the weights can meaningfully fluctuate across the different experts, favoring those that perform best at different times.

\subsection{Experiments in iid setting}
We start by examining our proposed methods, in a scenario where data are iid both in a classification and regression task. Additional experiments are reported in Appendix~\ref{sec:app_addexp}.

\subsubsection{Regression in iid setting}
To examine the practical behavior of COMA, we apply it to a real-world dataset. Specifically, the dataset comprises $75\,000$ observations pertaining to Airbnb apartments in New York City \citep{bernardi2023determinants}. The response variable is represented by the logarithm of the nightly price, while the covariates include information about the apartment and its geographical location. The data has been partitioned into 75 rounds, each consisting of $1000$ observations and a single test point. At each iteration, $K$ split conformal prediction intervals are constructed based on the data observed at round $t$ using various models. In the first simulation, $K=4$ regression algorithms are employed: linear model, lasso regression with penalty parameter set to 0.1, ridge regression with penalty parameter set to 0.1, and random forest with 250 trees. In the second scenario, a neural net with 8 nodes is added, and so $K=5$. In both cases, the random forest emerges as the best model, producing narrower intervals. However, whereas the differences with other models are substantial in the first scenario, in the second scenario the neural net also delivers satisfactory results.

\begin{figure}
    \centering
    \includegraphics[width=0.95\textwidth]{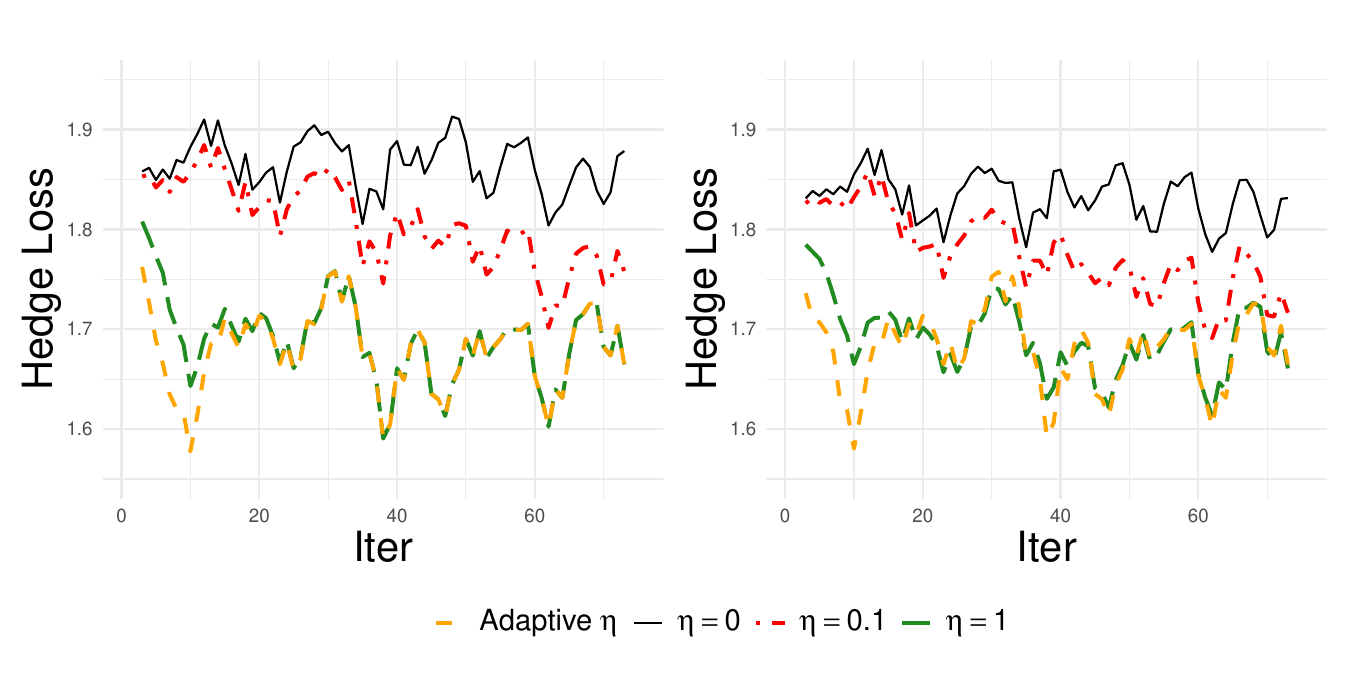}
    \caption{Hedge loss ($h_t$) obtained during various iterations with either a constant or adaptive learning rate scheme. The case $\eta = 0$ coincides with the standard (non-weighted) majority vote. The case with $K=4$ algorithms is shown in the left plot, while the case with $K=5$ algorithms is shown in the right plot. The series have been smoothed using a moving average $(5, \frac{1}{5})$. In both cases, COMA with adaptive $\eta$ quickly achieves the smallest loss}
    \label{fig:ewa_bnb}
\end{figure}

\begin{figure}
    \centering
    \includegraphics[width=1\textwidth]{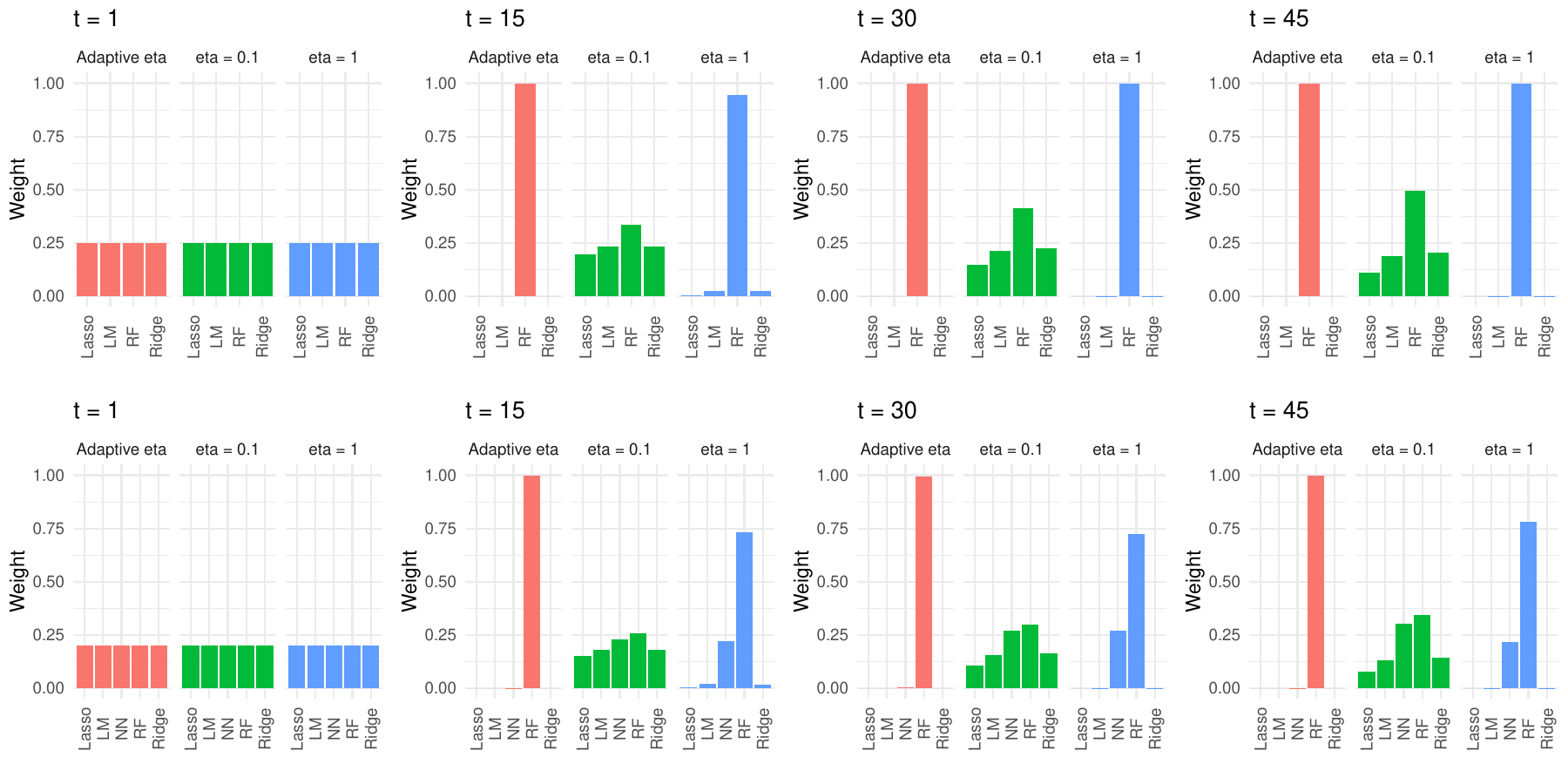}
    \caption{Weights assumed by the regression algorithms during different iterations. The case with $K=4$ algorithms is shown in the top row, while the case with $K=5$ algorithms is shown in the bottom row. After few iterations, the strategy with adaptive $\eta$ assigns full weight to the random forest. The COMA method with fixed $\eta$ is more conservative in assigning weights}
    \label{fig:weights_airbnb_t}
\end{figure}

The weights are adjusted using the COMA method with either adaptive or fixed values of $\eta$. For Algorithm~\ref{alg:ewa}, when $\eta^{(t)}=\eta$ is kept constant across iterations, two values are considered: $\eta=0.1$ and $\eta=1$. The case $\eta=0$ is included as a benchmark; it essentially corresponds to the solution proposed by \cite{cherubin2019}, where the weights remain uniform and fixed over time.
In both cases, the loss of the COMA algorithm with adaptive $\eta$ decreases rapidly and then stabilizes (Figure \ref{fig:ewa_bnb}). This behavior is explained by the fact that the algorithm quickly assigns unit mass to the most efficient model that is the random forest (Figure~\ref{fig:weights_airbnb_t}). In contrast, when $\eta$ is set at 0.1, the decrease in loss is slower, and the algorithm behaves more conservatively in assigning weights. Finally, when the weights remain constant throughout the iterations, the loss stays stable and no improvements are observed.


 {We also compare our approach with several alternative methods. As a first benchmark, we apply conformal prediction to an ensemble model constructed from the $K=5$ algorithms used in our analysis, where predictions are aggregated using a simple average. Although this approach guarantees valid coverage, it requires access to the estimated outputs of all $K$ models to construct the prediction set. In contrast, our method relies only on the $K$ sets themselves and can thus be applied in decentralized scenarios where the different experts cannot communicate. A second competitor is the naive union of the $K$ sets. This strategy also guarantees valid coverage, but typically results in excessively wide prediction sets (in terms of size, this can be interpreted as a worst-case benchmark). On the other hand, by the Bonferroni inequality, the intersection of the $K$ sets ensures a coverage of at least $1-K\alpha$, which may become meaningless as $K$ grows. To provide a fair comparison, we obtain the conformal prediction sets at level $2\alpha/K$ using the $K=5$ different algorithm. Under this choice, the intersection achieves a coverage of at least $1-2\alpha$ (as our method). As a final benchmark, we consider standard linear regression to construct a prediction set at level $1-\alpha$. Although this method does not guarantee validity due to the potential misspecification of the underlying assumptions, it is often considered robust in practice. The miscoverage level $\alpha$ is set to $0.05$.}

\begin{table}[]
    \centering
    \begin{tabular}{c|cccccc}
    \hline
        Method & Ensemble & \makecell{Linear\\ Regression} & Bonf. & Union & \makecell{Adapt. COMA\\ ($K=4$)} & \makecell{Adapt. COMA\\ ($K=5$)}\\
        \hline
        Size & 1.738 & 1.874 & 1.863 & 2.229 & 1.665 & 1.670\\
        Coverage & 0.973 & 0.987 & 0.987 & 0.987 & 0.973 & 0.973\\
    \hline
    \end{tabular}
    \caption{Comparison of empirical size and coverage with $\alpha = 0.05$. The best method in terms of efficiency is COMA with adaptive $\eta$}
    \label{tab:tab_bench}
\end{table}

 {As reported in Table~\ref{tab:tab_bench}, the union method yields excessively large sets, whereas the ensemble provides better results. Overall, COMA with adaptive stepsize outperforms the competing methods while maintaining comparable coverage. In general, COMA exhibits strong empirical performance, while still ensuring adequate empirical coverage. In particular, the adaptive–$\eta$ version is able to perform model selection in scenarios where one model clearly outperforms the others. For the COMA method, the observed output is always a non-empty interval. Assumption~\ref{ass:1} appears reasonable, as the maximum empirical counterpart of the total elementwise correlation in \eqref{eq:ass-neg-tot-corr} is $0.002$ for both the fixed-$\eta$ and adaptive-$\eta$ methods.}

\subsubsection{Classification in iid setting}
 {For the classification experiment, we generate a synthetic dataset with $2000$ observations as follows. 
We consider $p=8$ covariates sampled from a multivariate Gaussian distribution with mean zero and equicorrelation structure, with $\sigma_{ij}=0.5$ for $i \neq j$. The response $y \in \{1,\ldots,8\}$ is then drawn from a multinomial distribution with probabilities proportional to $w_j(x) \propto \exp\{x^\top \beta_j\}, j=1,\ldots,8,$ where each coefficient vector $\beta_j \in \mathbb{R}^p$ is independently sampled 
from a standard normal distribution. Specifically, following a burn-in phase of $1000$ iterations, we re-trained four different classifiers at each step, and the weights for these algorithms are adjusted using the COMA method. In particular, the four classifier chosen from the R package \texttt{mlr} are Linear Discriminant Analysis (LDA), Quadratic Discriminant Analysis (QDA), Random Forest and Neural Net (we refer to \citet{azzalini2012} for an introduction to the methods). At each time step, split conformal inference is used to obtain a prediction set for $y^{(t)}$ and the score used are $s(x,y)=1-\hat{\mu}(x;y_d)$ where $\hat{\mu}(x;y_d)$ is the predicted probability for the label $y_d\in\mathcal{Y}=\{1,\dots,8\}$.}

\begin{table}
    \centering
    \begin{tabular}{c|cccccccccc}
    \hline
        Method & NN & RF & QDA & LDA & \makecell{COMA\\(Adap. $\eta$)} & \makecell{COMA\\($\eta=0.1$)} & Union & Bonf. & Ensemble \\
        \hline
        Size & 3.056 & 3.106 & 2.897 & 2.488 & 2.489 & 2.476 & 4.159 & 2.647 & 2.619\\
        Coverage & 0.902 & 0.903 & 0.889 & 0.899 & 0.896 & 0.890 & 0.965 & 0.897 &  0.897 \\
        \hline
    \end{tabular}
    \caption{Comparison of empirical size and coverage at significance level $\alpha = 0.1$ between the underlying algorithms and COMA (with both adaptive and fixed $\eta$). For reference, we also include the union, the intersection, and the sets obtained via conformal prediction using an ensemble of the base algorithms as  predictive model }
    \label{tab:classiid}
\end{table}

\begin{figure}
    \centering
    \includegraphics[width=\linewidth]{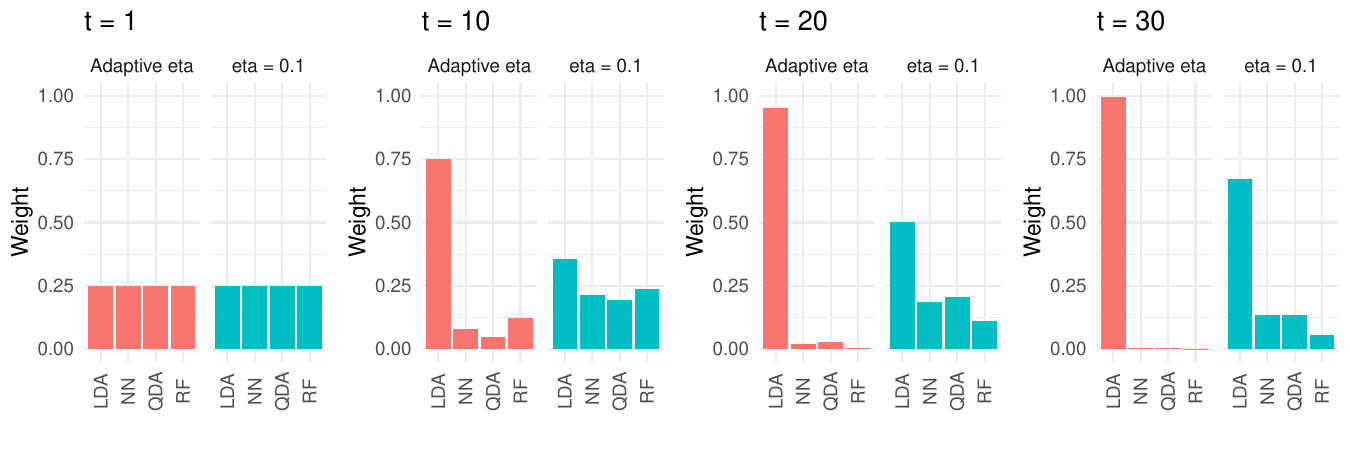}
    \caption{Weights assumed by the classification algorithms during different iterations. After few iteration the strategy with adaptive $\eta$ puts unit mass on LDA, the best-performing algorithm}
    \label{fig:classiidw}
\end{figure}

\begin{figure}
    \centering
    \includegraphics[width=.85\linewidth]{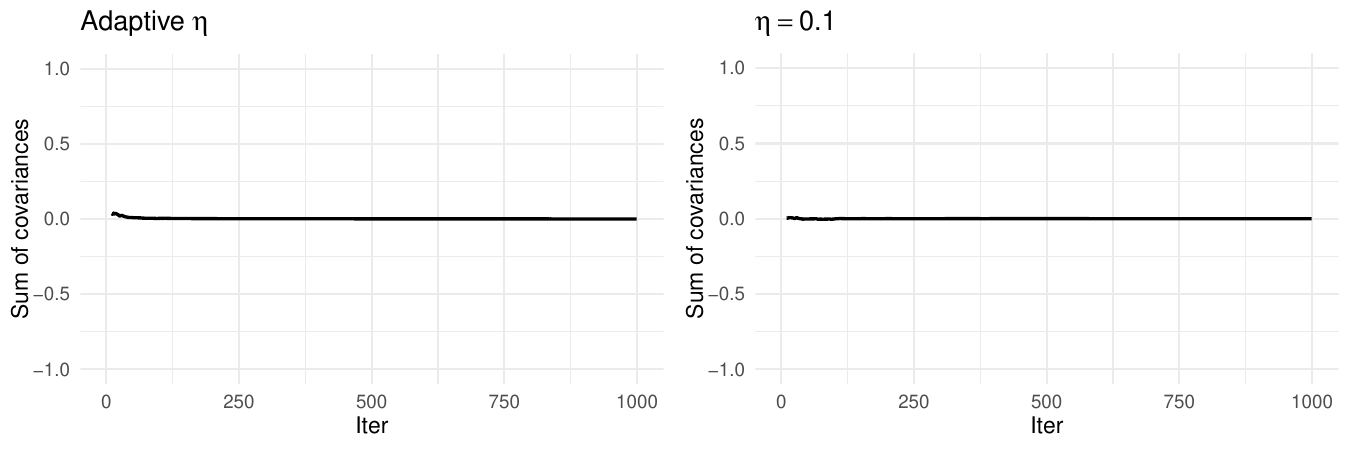}
    \caption{Empirical total elementwise correlation, computed as the empirical sum over time $(T)$ of the covariances between $\{w_{k}^{(t)}\}_{t=1}^T$ and $\{\phi_{k}^{(t)}\}_{t=1}^T$. Both series remain near zero over all iterations}
    \label{fig:iid_scvs}
\end{figure}

 {We apply our COMA procedure with both adaptive and fixed $\eta$ (set to $0.1$). For comparison, we also include, as before, the union of the label sets and the intersection obtained using each method trained at a miscoverage level of $2\alpha/K$. In addition, we apply conformal prediction using an ensemble of base models. From Table~\ref{tab:classiid}, our methods achieve the smallest set sizes while maintaining empirical coverage close to $1-\alpha$. The factor of $2$ does not appear to be present in the miscoverage, as the miscoverage rate remains close to $\alpha$. The same behavior is observed when applying the intersection of the sets (Bonferroni). This can be explained by the fact that Bonferroni’s inequality is tight near-independence, whereas in this case there is strong dependence, since all methods are trained on the same data. Figure~\ref{fig:classiidw} shows that the weights quickly concentrate on LDA, which is the best-performing model. Moreover, the empirical counterpart of the quantity defined in \eqref{eq:ass-neg-tot-corr} remains close to zero throughout all iterations, as illustrated in Figure~\ref{fig:iid_scvs}. }


\subsection{Experiments in non-iid setting}
We now evaluate our COMA procedure as a wrapper for quantile tracking and ACI through some experiments on real and simulated datasets. In particular, we empirically evaluate the methods proposed both in Section~\ref{subsec:wrapper-aci} and in Section~\ref{subsec:aci-2}.

\subsubsection{ELEC2 dataset}\label{sec:elec2}
We now use the algorithm described in Section~\ref{subsec:wrapper-aci} in a real-world application. We define $K=3$ different regression algorithms, and we employ quantile tracking with a decaying step size $\gamma^{(t)} \propto t^{-1/2-\epsilon}$. The score function corresponds to absolute residuals. In particular, we use the \texttt{ELEC2} \citep{harries1999} data set that monitors electricity consumption and pricing in the states of New South Wales and Victoria in Australia, with data recorded every 30 minutes over a 2.5-year period from 1996 to 1999. For our experiment, we utilize four covariates: \texttt{nswprice} and \texttt{vicprice}, representing the electricity prices in each respective state, and \texttt{nswdemand} and \texttt{vicdemand}, denoting the usage demand in each state. The response variable is \texttt{transfer}, indicating the quantity of electricity transferred between the two states. We narrow our focus to a subset of the data, retaining only observations within the time range of 9 am to 12 pm to mitigate daily fluctuation effects.
The models used correspond to three distinct linear models based on different covariates. The first model relies on the entire set of covariates at time $t$, the second model employs $y^{(t-1)}$ and $y^{(t-2)}$ as covariates (it corresponds to an AR(2)), while the third model supplements the latter with the covariate \texttt{nswprice}.

\begin{figure}
    \centering
    \includegraphics[width=1\textwidth]{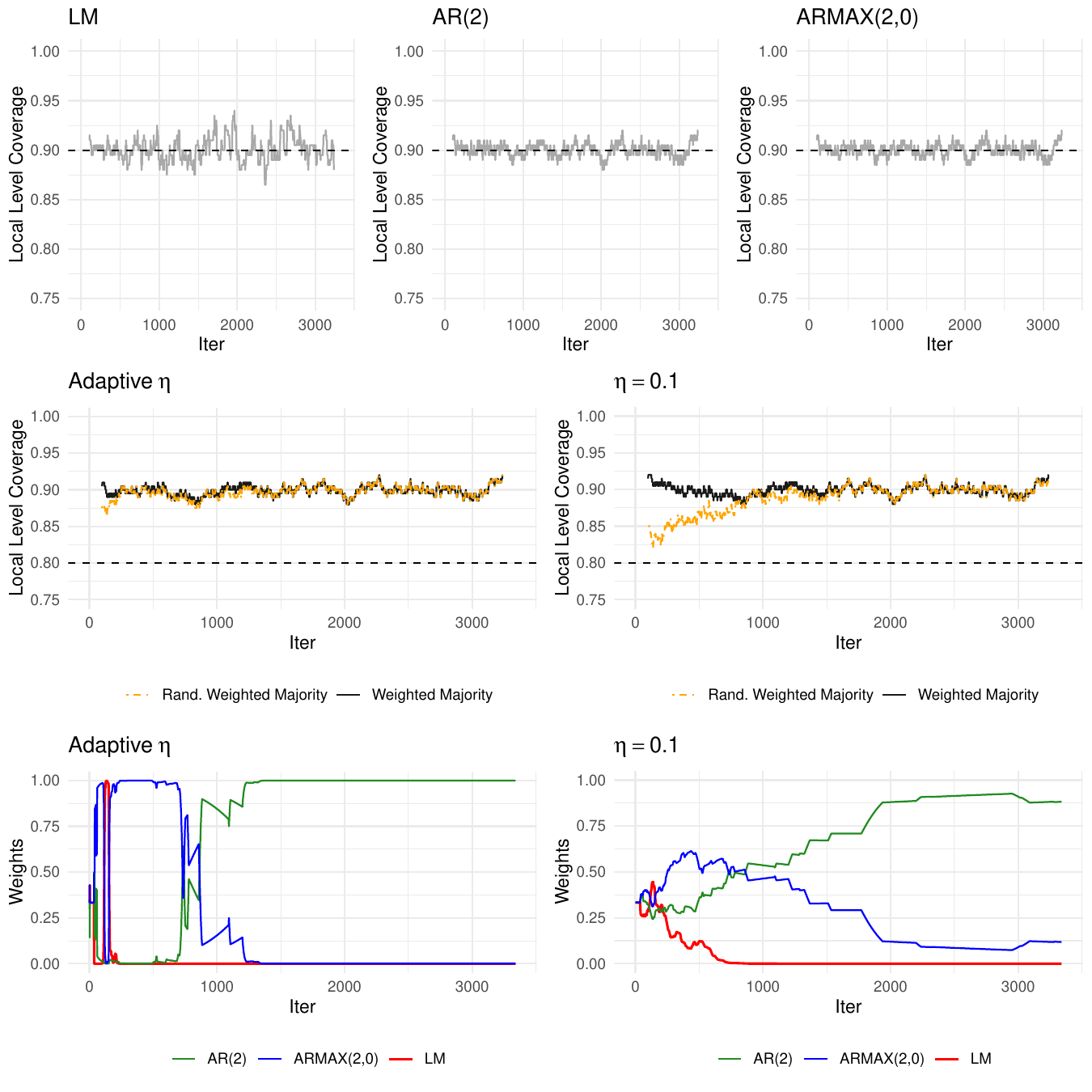}
    \caption{First row: Local level coverage for the quantile tracking using the three regression models. The dashed line represents the level $1-\alpha$. Second row: Local level coverage for the COMA method with adaptive $\eta$ or $\eta=0.1$.  The dashed line represents the level $1-2\alpha$. 
    Third row: Weights obtained by the various model for the decentralized COMA algorithm applied in \texttt{ELEC2} dataset. The strategy with adaptive $\eta^{(t)}$ is more aggressive than the one with $\eta^{(t)}=\eta$ in assigning weights and it performs model selection in most of the iterations
    }
    \label{fig:elec2_iter}
\end{figure}

The coverage performance of the methods is compared using the local level coverage on $200$ data points:
\begin{equation}
    \label{eq:llc}
    \textrm{localCov}^{(t)} := \frac{1}{200}\sum_{i=t-100+1}^{t+100} \ind\left\{y^{(i)} \in \setc_k^{(i)}\left(q_k^{(i)}\right)\right\},
\end{equation}
where $\setc_k^{(t)}(q_k^{(t)})$ is the set obtained by the $k$-th regression algorithms with the quantile fixed at $q_k^{(t)}$. The local coverage is also used to measure the performance of our merging procedure. In particular, we compare the use of an adaptive learning parameter and a small fixed $\eta=0.1$. As can be seen in Figure~\ref{fig:elec2_iter}, the three base models and the adaptive COMA procedure obtain local level coverage that oscillates around the prespecified level $1-\alpha$. This is because the weights concentrate on one model over time; in particular, the chosen model is not always the same, as can be seen in the last row in Figure~\ref{fig:elec2_iter}. The procedure proves appealing in this setting: when a clear winner emerges, it concentrates the weights on the best model while, in the absence of dominance, it allocates them across the most appropriate models while maintaining coverage. 

\begin{table}
\begin{tabular}{c|cccccccc}
  \hline
 & LM & AR & ARMAX & Adapt. $\eta$ & \thead{Adapt. $\eta$ \\+ Rand.} & $\eta=0.01 $ & \thead{$\eta=0.01$\\ + Rand.} & $\eta=0$ \\ 
  \hline
$L^{(T)}$ & 1288.76 & 787.04 & 807.16 & 788.20 & 780.34 & 789.51 & 745.53 & 808.24 \\ 
  Coverage & 0.900 & 0.902 & 0.902 & 0.900 & 0.897 & 0.901 & 0.889 & 0.905 \\ 
   \hline
\end{tabular}
\caption{Cumulative loss ($L^{(T)}$) and empirical coverage obtained by the various methods. The losses of the proposed methods are better (or similar) than the cumulative loss obtained by the best algorithm. The column $\eta=0$ represents a baseline where weights are not updated }
\label{tab:res1}
\end{table}

In Table~\ref{tab:res1}, we compare the cumulative loss and the empirical coverage obtained by the various methods proposed with and without the use of randomization. In addition, the case $\eta=0$ corresponds to the majority vote with uniform weights. As can be seen, the cumulative loss of our proposed methods (with fixed and adaptive $\eta$) is better than the standard majority vote approach in all cases. The best result in terms of cumulative loss, in this case, is obtained by fixing $\eta=0.01$ and adding randomization. This is due to the fact that during some iterations the two best models have a similar weight and their sum is close to one. In this situation, the majority vote reduces to the intersection of the two sets most of the time. As an example, suppose that $w^{(t)} \approx (0.6, 0.4, 0)$ and $u^{(t)} \approx 1/2$, then the threshold is $3/4$ and in this case the majority vote set must contain the points of the first two sets (the intersection of $\setc_1^{(t)}$ and $\setc_2^{(t)}$). Indeed, during the first iterations, the local level coverage for COMA with fixed $\eta$ and randomization is observed to be closer to $1-2\alpha$. The cumulative loss of the methods using randomization is smaller than the cumulative loss incurred by the best method, which in this case is AR(2). In particular, the factor of 2 in \eqref{eq:regret-bound} does not appear. The empirical coverage is approximately $0.9$ in all scenarios. The maximum total elementwise correlation between the weight and error vectors defined in \eqref{eq:ass1} is $0.02$, suggesting that the assumption of negative total elementwise empirical correlation appears to hold. The proportion of empty sets produced by both methods is below 1\%.

\subsubsection{Google stock prices}\label{sec:google}
\begin{figure}[H]
    \centering
    \includegraphics[width=1\textwidth]{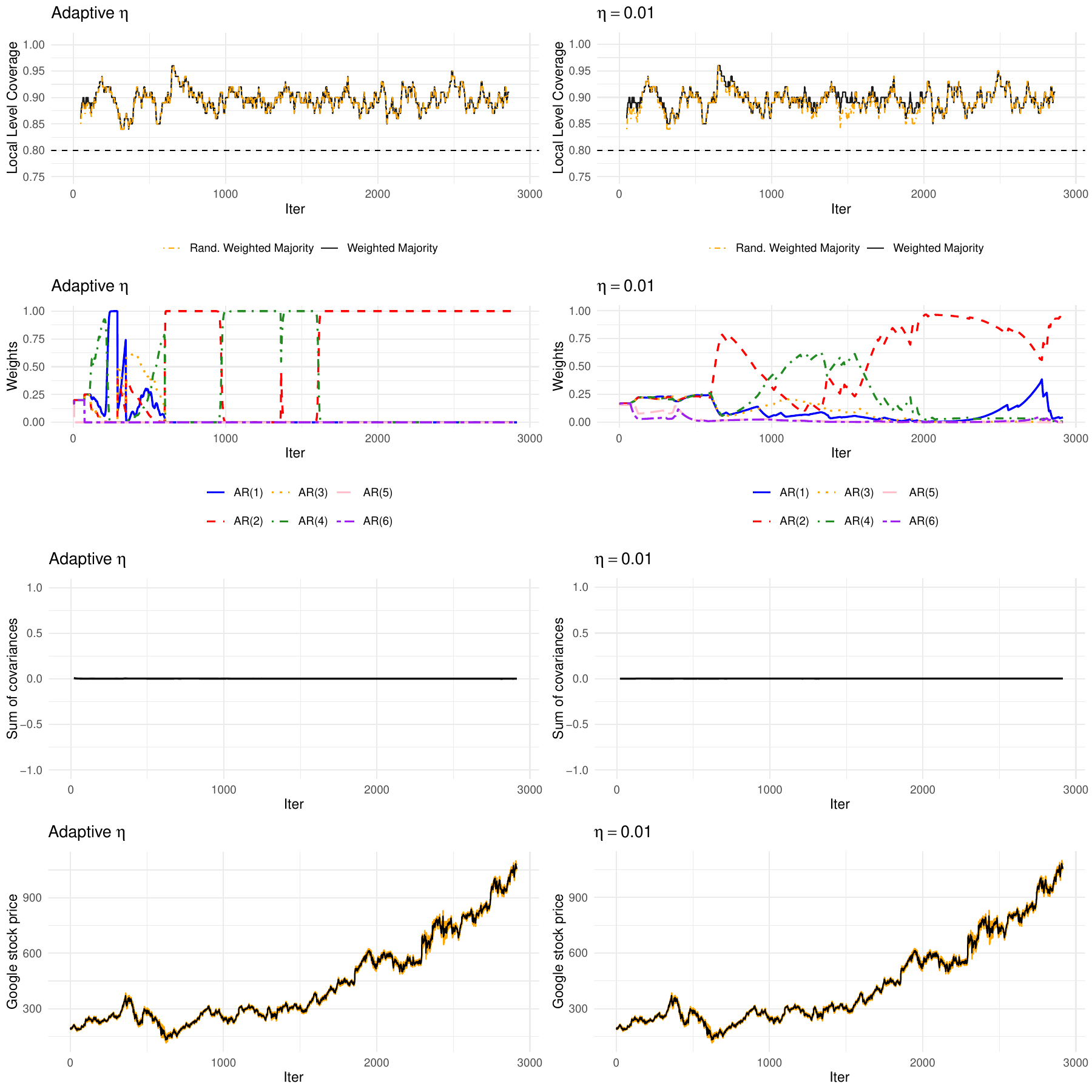}
    \caption{Local level coverage obtained in a window containing 100 data points (first row), weights assumed during the different iterations by the various models (second row), empirical sum of the quantity $\frac{1}{T}\sum_{t=1}^T\sum_{k=1}^K(\phi_k^{(t)}-\bar{\phi}_k)(w_k^{(t)}-\bar{w}_k)$ obtained using an increasing number of observations (third row) and series of the original stock prices with corresponding prediction intervals (fourth row). The black dashed lines in the first row represent the level $1-2\alpha$. The results corresponding to the strategy with adaptive $\eta$ are presented in the first column, while the results corresponding to the strategy with $\eta=0.01$ are presented in the second column}
    \label{fig:googl}
\end{figure}

Inspired by \citet{angelopoulos2023pid}, we analyze the historical series comprising the single stock open prices of Google from January 1, 2006, to December 31, 2014 \citep{nguyen2018stock}. The compared models consist of six different autoregressive models incorporating temporal lags from 1 to 6, trained on the logarithmic scale of the opening price, while the forecasts are obtained on the original scale. The series exhibits nonstationary behavior, as can be observed in the last row in Figure~\ref{fig:googl}. To construct the merged set, we apply the algorithm introduced in Section~\ref{subsec:wrapper-aci}, setting $\alpha = 0.1$. A similar analysis is presented in~\citet{angelopoulos2023pid}, which considers a single autoregressive model with three lags. The parameter $\gamma$ is updated as $\gamma^{(t)}=0.1B^{(t)}$, where $B^{(t)}$ is the maximum observed score in a trailing window of 100 iterations, and the models are re-trained at every time step. The score function corresponds to absolute residuals $s(x^{(t)}, y^{(t)}) = |y^{(t)} - \hat{\mu}_k^{(t)}(x^{(t)})|,\, k \in [K],$ and prediction sets are obtained as in \eqref{eq:qt_set}. The weights are updated using an adaptive $\eta$ or fixing $\eta$ to the level $0.01$, this small value of $\eta$ allows the mixing of the different intervals during iterations.


It appears that autoregressive models with two and four lags perform better than the others. Particularly, in the case of adaptive $\eta$, after the first iterations, they alternate during different times (Figure~\ref{fig:googl}). As expected, the strategy with a small fixed $\eta$ is more conservative in assigning weights and continues to assemble the different models during all iterations. In both methods, with adaptive $\eta$ and fixed $\eta$, local coverage remains around the level of $1-\alpha$. The quantity $\frac1T \sum_{t=1}^T \sum_{k=1}^K (\phi_k^{(t)} - \Bar{\phi}_k)(w_k^{(t)} - \Bar{w}_k)$ is around zero during all iterations.

To summarize the results, it seems that our proposed method is able to adapt to the changes of the distributions observed in the data. The total negative empirical correlation assumption seems to be reasonable in both cases.

\subsubsection{Additional simulations on quantile tracking}
\label{sec:aci_additional}
We investigate the quantile tracking method in conjunction with ``decentralized'' COMA with adaptive $\eta$ (Section~\ref{subsec:wrapper-aci}). We conducted a case study in which changes in the marginal distribution also impact the weights of the two algorithms. In this case, we use simulated data and we let the distribution vary during the iterations as described below. We suppose to have two covariates, $x_1$ and $x_2$, and the data are generated in blocks, with only one of the two covariates influencing the response at a time. Specifically, at each time $t$ we have $y^{(t)} = 2x^{(t)} + \varepsilon^{(t)}$ where 
\[
x^{(t)} = 
\begin{cases}
x_1^{(t)}, \quad t \in \{[0,50) \cup [100,150) \cup [250,350) \cup [450,550) \cup \dots \cup [1250, 1350)\}\\
x_2^{(t)}, \quad t \in \{[50,100) \cup [150,250) \cup [350,450) \cup \dots \cup [1350, 1450]\}
\end{cases}
\]
and $x_1^{(t)}, x_2^{(t)},\varepsilon^{(t)}$ are generated independently from a standard normal distribution. Two standard linear models are used, one using $x_1$ as a regressor and the other using $x_2$. Prediction intervals are generated using the quantile tracking method with the absolute residual as the score function. The models are re-trained at every time step using the most recent $100$ observations. The parameter $\gamma$ is set to $1$ and the entire procedure is repeated $1000$ times.

\begin{figure}[H]
    \centering
    \includegraphics[width=.65\textwidth]{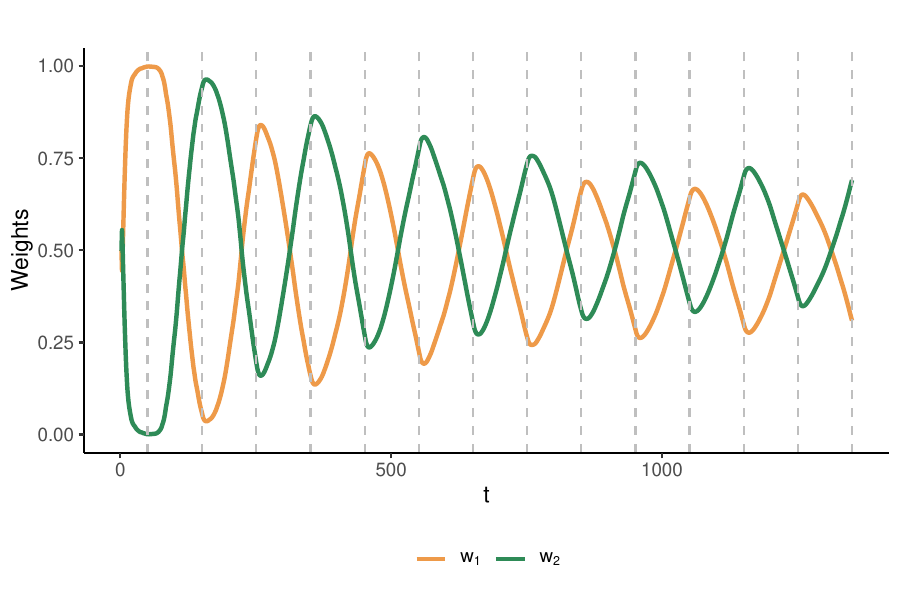}
    \caption{Average weights across 1000 iterations of the linear model, using either $x_1$ or $x_2$ as the regressor. The dotted lines indicate the time points at which the distribution changes}
    \label{fig:wt_aci}
\end{figure}

In the case where only two models are utilized, the majority vote (without randomization) is reduced to selecting the model with the higher weight (it performs model selection by definition). From Figure~\ref{fig:wt_aci}, the COMA method alternates the weights based on changes in the marginal distribution.  In other words, the method can adapt to the changes and perform model selection. In particular, within each time window, the selected model switches between the two models around the midpoint of the window. The less aggressive behavior in assigning weights is due to the fact that $\eta^{(t)}$ decreases over time. The empirical miscoverage over the 1000 replications of our COMA procedure is 0.113; while the miscoverage rates for the two linear models are respectively 0.101 and 0.102. The maximum empirical correlation between the weights and the errors across the $1000$ replications is $0.167$.

\subsubsection{Classification under distribution shift}
 {We consider a classification example in a decentralized setting, conducted under a non-iid scenario. The dataset consists of hourly air quality measurements collected in the city of Chanping   \citep{beijing}. The response variable corresponds to the levels of PM10 concentration, discretized into five classes of equal cardinality, while the covariates comprise meteorological indicators and the concentrations of other atmospheric pollutants. The models used include a Random Forest with 150 trees, a neural network with 4 nodes, a neural network with 6 nodes, and Linear Discriminant Analysis (LDA). For this experiment, we employed the ACI method with the parameter $\gamma$ set to 0.01. Since we are dealing with a classification problem, the loss is bounded and therefore cannot take infinite values (also if $\alpha^{(t)}<0$). The significance level $\alpha$ is set to 0.1.}

 {Also in this case, where the loss takes only integer values, the sets obtained via our COMA procedure attain a loss that is comparable to, and in some instances even better than, that of the best model. Moreover, the empirical coverage remains close to the nominal level $1-\alpha$ (see Table~\ref{tab:clasnoiid}). The quantity in \eqref{eq:ass1} remains near zero across all iterations, with a maximum of 0.012 for COMA with adaptive $\eta$ and 0.006 for COMA with fixed $\eta$. }

\begin{table}[]
    \centering
    \begin{tabular}{c|ccccccccc}
    \hline
         & RF & \makecell{NN\\(4)}& \makecell{NN\\(6)} & LDA & Adapt. $\eta$ & \thead{Adapt. $\eta$ \\+ Rand.} & $\eta=0.01 $ & \thead{$\eta=0.01$\\ + Rand.} & $\eta=0$\\
         \hline
       $L^{(T)}$  & 1962 & 2355 & 2267 & 2414 & 1964 & 1959 & 1967 & 1945 & 2016 \\
       Coverage & 0.895 & 0.894 & 0.895 & 0.894 & 0.896 & 0.895 & 0.896 & 0.890 & 0.891\\
       \hline
    \end{tabular}
    \caption{Cumulative loss ($L^{(T)}$) and empirical coverage obtained by the various methods. The column $\eta=0$ represents a baseline where weights are not updated}
    \label{tab:clasnoiid}
\end{table}

\subsubsection{ELEC2 dataset --- centralized setting}
We conclude the experimental section with a study addressing the \emph{centralized} case described in Section~\ref{subsec:aci-2}. We applied the procedure to the ELEC2 dataset, fixing the target coverage level at $0.9$ and adopting the loss function $\arctan(\ell^{(t)})$. 
The use of a bounded loss function is necessary, since ACI can output $\mathbb{R}$ as an interval. 
We consider both the adaptive $\eta$ strategy and a fixed $\eta = 0.1$, using the three regression algorithms described in Section~\ref{sec:elec2}. 
In addition, we apply ACI with a regression model obtained by ensmebling the three algorithms by a simple average. 
The learning rate of the ACI algorithm is set to $0.05$ for all methods, and we use~\eqref{eq:llc} to compare their coverage performance. 

\begin{figure}
    \centering
    \includegraphics[width=1\textwidth]{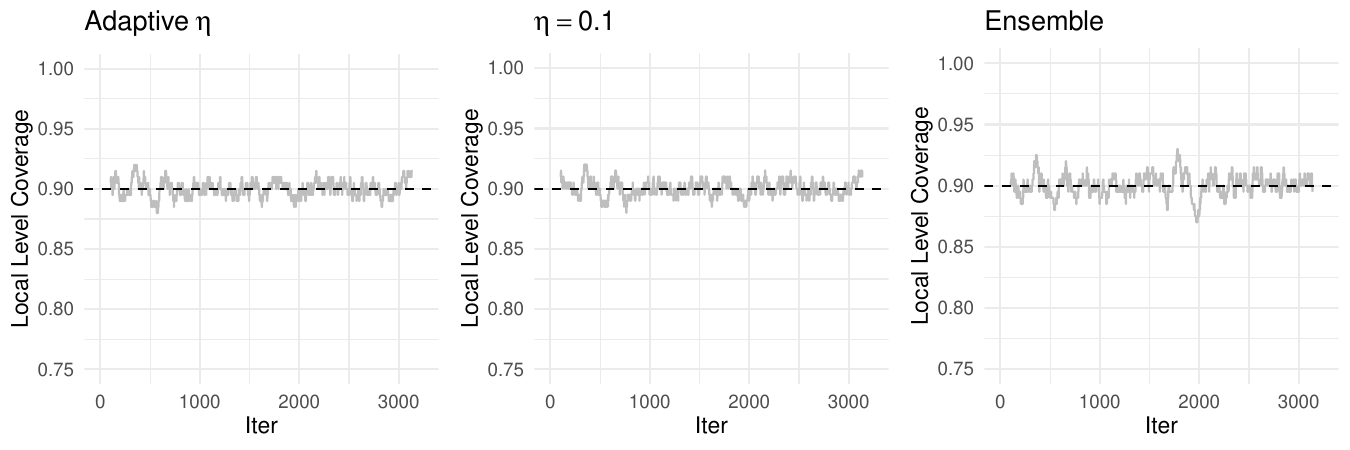}
    \caption{Local level coverage for \emph{direct} ACI (Section~\ref{subsec:aci-2}) for COMA with adaptive $\eta$ and $\eta=0.1$ applied to \texttt{ELEC2} dataset. The third column represents ACI directly applied to an ensemble of the three base models. The dashed lines represent the target level $1-\alpha$
    }
    \label{fig:ewma+aci}
\end{figure}

The results, reported in Figure~\ref{fig:ewma+aci}, show that the local coverage oscillates around the target level $1-\alpha$ for all three methods. 
The methods yield comparable results in terms of empirical size and in the proportion of cases where the sets coincide with $\mathbb{R}$ (i.e., when $\alpha \le 0$). 
However, our strategy with adaptive $\eta$ achieves the best performance in terms of set size (Table \ref{tab:ewdirec}), whereas the strategy with fixed $\eta$ has the smallest proportion of cases where the sets coincide with $\mathbb{R}$.

\begin{table}[]
    \centering
    \begin{tabular}{c|rrr}
    \hline
    & Adaptive $\eta$ & $\eta = 0.1$ & Ensemble \\
    \hline
       Empirical size & 0.230 & 0.234 & 0.238 \\  
       \% set$=\mathcal{Y}$ & 0.926 & 0.556 & 1.575 \\ 
    \hline
    \end{tabular}
    \caption{Empirical size and percentage of times the sets equal $\mathcal{Y}$. The strategy with adaptive $\eta$ achieves the best performance in terms of empirical size, while the proportion of cases where the methods produce sets of infinite size is negligible}
    \label{tab:ewdirec}
\end{table}

\section{Conclusion}\label{sec:conclusion}
Our work presents a method for aggregating conformal prediction sets derived from different prediction algorithms. Specifically, a method based on majority vote is proposed in which the weight of each individual prediction model is updated based on its past performance.  {The proposed method requires access only to the prediction sets, which can be particularly advantageous in decentralized settings and is computationally efficient. Our approach can be seen as a principled wrapper around conformal prediction, specifically designed for streaming and online settings. }

In some cases, when one algorithm outperforms the others, all the mass is assigned to the best-performing model, effectively resulting in model selection. In other cases, the weight shifts between models, depending on which one performs better at a given time.
Particularly, in the case of iid data, it is very common for there to exist a \emph{best} model that outperforms the others in terms of size of the outputted sets.

It is feasible to establish an upper bound for the loss incurred by our method relative to the loss of the best-performing algorithm, translating into a form of regret for our method.
We also tested our method in cases where the data underwent a distribution shift, on top of the quantile tracking and ACI methods. 

One limitation is that in practice we often see $1-\alpha$ coverage, even when the current theory only predicts it to be at least $1-2\alpha$. Another is that we assumed that we get to observe the true label in every round; extending it to intermittent or delayed feedback would be interesting future work.

\subsection*{Acknowledgments}
AR acknowledges funding from NSF grant IIS-2229881, and thanks Rina Foygel Barber, Ryan Tibshirani and Emmanuel Candès for useful conversations around model selection in conformal prediction.

\bibliography{biblio.bib}

\appendix
\section{Conformal inference and adaptive conformal inference}
\label{sec:app_ci}
In this section we briefly introduce split (or inductive) conformal prediction \citep{papadopoulos2002, lei2018}. A more detailed introduction can be found in \citet{angelopoulos2021}. Suppose that we have a fitted prediction model $\hat{\mu}$ used to predict the value of $Y^{(t)}$ given $X^{(t)}=x^{(t)}$. In addition, let us assume that we have a calibration set $\mathcal{D}_{cal} = (x^{(i)}, y^{(i)})_{i=1}^{t-1}$ that is different from the data where we trained our prediction model. In this scenario, we introduce $s: \mathcal{X} \times \mathcal{Y} \to \mathbb{R}$ as the \emph{conformal score} function, assessing the degree of agreement between the value $y$ and the predictions from the model. A commonly used score function in regression analysis is the absolute residual, expressed as
\[
s(x,y) = | y - \hat{\mu}(x) |,
\]
where $\hat{\mu}(x)$ represents the point prediction of the model $\hat{\mu}$ using the predictors in $x$. We can now define the fitted quantiles of the scores in the calibration set as
\[
q(\beta) := \inf\left\{b: \frac{1}{|\mathcal{D}_{cal}|} \sum_{(x^{(i)}, y^{(i)}) \in \mathcal{D}_{cal}} \ind\{s(x^{(i)}, y^{(i)}) \leq b\} \geq \beta \right\}.
\]
The conformal prediction set for $Y^{(t)}$ based on $X^{(t)}=x^{(t)}$ with error level $\alpha\in(0,1)$ is given by 
\begin{equation}\label{eq:conf_set}
    \setc^{(t)} := \setc^{(t)}(\alpha) = \{y \in \mathcal{Y}: s(x^{(t)}, y) \leq q(1-\alpha)\}.
\end{equation}
If data are iid (or at least exchangeable), then it is possible to prove that $\pr(Y^{(t)} \notin \setc^{(t)})\leq \alpha$.

The results can also be extended for classification problems; in this case, we only need different conformity scores functions (since, for example, residual based scores are generally not appropriate). One possible solution is to use $s(x,y)=1-\hat{\mu}(x;y_d)$, where $\hat{\mu}(x;y_d)$ is the predicted probability for the label $y_d\in\mathcal{Y}=\{y_1,\dots,y_D\}$ given the value of $x$. Other proposals are obtained in \citet{romano2020}.

In the case of adaptive conformal inference \citep{gibbs2021}, the error level varies over time, as described in \eqref{eq:alpha_up}. However, the prediction sets at each time $t$ are obtained as in \eqref{eq:conf_set}, where $\mathcal{D}_{cal}$ is the calibration set that contains observations before time $t$ that were not used to train the prediction model. In the described scenario, it is not guaranteed that the level $\alpha^{(t)}$ lies in the interval $(0,1)$. Specifically, when $\alpha^{(t)} \leq 0$, then $\setc^{(t)}(\alpha^{(t)})=\mathcal{Y}$, while when $\alpha^{(t)}\geq 1$, we have $\setc^{(t)}(\alpha^{(t)})=\emptyset$.

\section{Additional experiments}\label{sec:app_addexp}
\subsection{Additional experiments on real datasets}
{We evaluated our COMA method, described in Section~\ref{sec:dynamic}, on multiple real datasets to assess its performance in terms of coverage and size. To construct the sets, we used split conformal prediction with absolute residuals as conformity scores. We considered two scenarios with $K=6$ base models. In the first scenario, the models used are: a linear regression, a lasso regression with regularization parameter $0.1$, a ridge regression with regularization parameter $0.1$, a random forest with $250$ trees, a neural network with $4$ nodes, and a neural network with $8$ nodes. In the second case, the models are three neural nets with $4,6,8$ nodes and three random forests with $50,150,250$ trees.}

{As benchmarks, we included prediction intervals from standard linear regression, split conformal prediction with an ensemble of the six models as base predictor (the ensemble is obtained by using a simple average), the union of the sets obtained from split conformal prediction, and the intersection of the sets obtained from split conformal prediction calibrated at level $\alpha' = 2\alpha/K$. For each dataset, we sampled $500$ observations and applied the aforementioned methods using data from the $250$-th observation onward. At each iteration, all models are retrained, and the results are reported in Table~\ref{tab:datasets} and Table~\ref{tab:datasets1}. In Tables~\ref{tab:datasets2} and \ref{tab:datasets4}, we report the total elementwise correlation defined in \eqref{eq:ass-neg-tot-corr}. The empirical values are close to zero across all experiments, suggesting that Assumption~\ref{ass:1} appears to hold. Our COMA method never produces empty sets in the experiments and overall is the best method. The performance of COMA with fixed $\eta$ and COMA with adaptive $\eta$ can differ, and in some cases the fixed-$\eta$ version outperforms the adaptive version. An explanation is that, after a few iterations, the adaptive-$\eta$ version concentrates all the weight on the best model, whereas the fixed-$\eta$ version distributes the weight across two or three algorithms. As a result, the randomized majority vote among two or three methods may align with the intersection, as discussed in the text.
}

\begin{table}
    \centering
    \begin{tabular}{c|rrrrrr}
    \hline
         Dataset & \makecell{COMA\\Adapt. $\eta$} & \makecell{COMA\\ $\eta=1$} & Ensemble & \makecell{Linear\\Regression} & Union & Bonf.  \\
         \hline
         Crime & 0.447 & 0.427 & 0.509 & 0.490 & 0.822 & 0.610\\
         \citep{communities_and_crime_183}& 0.900 & 0.888 & 0.928 & 0.972 & 0.984 & 0.940 \\
         \hline
         Aquatic toxicity & 0.990 & 0.956 & 0.974 & 1.040 & 1.290 & 1.266 \\
         \citep{qsar_aquatic_toxicity_505}& 0.900 & 0.896 & 0.940 & 0.944 & 0.972 & 0.960 \\
         \hline
         Diamonds & 0.810 & 0.782 & 0.887 & 1.115 & 1.252 & 0.928\\
         \citep{diamonds} & 0.892 & 0.884 & 0.908 & 0.900 & 0.968 & 0.936 \\ 
         \hline
         Concrete & 23.104 & 23.665 & 28.814 & 33.864 & 41.944 & 28.042\\
         \citep{concrete}& 0.916 & 0.924 & 0.932 & 0.888 & 0.988 & 0.940\\ 
         \hline
         Productivity & 0.487 & 0.477 & 0.524 & 0.484 & 0.672 & 0.633 \\
        \citep{productivity}& 0.944 & 0.932 & 0.912 & 0.860 & 0.960 & 0.944\\ 
         \hline         
    \end{tabular}
    \caption{Empirical results of the different methods across multiple datasets. The base models are: lasso regression, ridge regression, linear model, random forest and two different neural nets. For each dataset, the first row reports the empirical set size, while the second row reports the empirical coverage}
    \label{tab:datasets}
\end{table}

\begin{table}
    \centering
    \begin{tabular}{c|rrrrrr}
    \hline
         Dataset & \makecell{COMA\\Adapt. $\eta$} & \makecell{COMA\\ $\eta=1$} & Ensemble & \makecell{Linear\\Regression} & Union & Bonf.  \\
         \hline
         Crime & 0.444 & 0.413 & 0.483 & 0.490 & 0.615 & 0.606\\
         \citep{communities_and_crime_183}& 0.916 & 0.896 & 0.932 & 0.972 & 0.980 & 0.980\\ 
         \hline
         Aquatic toxicity & 0.910 & 0.876 & 0.980 & 1.040 & 1.260 & 1.238\\
         \citep{qsar_aquatic_toxicity_505}& 0.912 & 0.904 & 0.940 & 0.944 & 0.992 & 0.956\\ 
         \hline
         Diamonds & 0.791 & 0.768 & 0.834 & 1.115 & 1.058 & 0.896 \\
         \citep{diamonds} & 0.892 & 0.888 & 0.932 & 0.900 & 0.988 & 0.940 \\ 
         \hline
         Concrete & 21.535 & 23.999 & 24.968 & 33.864 & 35.670 & 26.208\\
         \citep{concrete}& 0.908 & 0.924 & 0.928 & 0.888 & 0.996 & 0.928\\ 
         \hline
         Productivity & 0.431 & 0.416 & 0.501 & 0.524 & 0.633 & 0.569\\
         \citep{productivity}& 0.896 & 0.888 & 0.912 & 0.880 & 0.972 & 0.940 \\
         \hline         
    \end{tabular}
    \caption{Empirical total elementwise correlation, defined in Assumption \ref{ass:1}, computed across different datasets. The base models are three random forests and three neural nets with different tuning parameters. For each dataset, the first row reports the empirical set size, while the second row reports the empirical coverage}
    \label{tab:datasets1}
\end{table}

\begin{table}[]
    \centering
    \begin{tabular}{c|rrrrr}
    \hline
        & Crime & \makecell{Aquatic\\ toxiticity} & Diamonds & Cement & Productivity\\
        \hline
        COMA (Adapt. $\eta$) & -0.001 & 0.000 & 0.000 & -0.003 & 0.001 \\
        COMA ($\eta$=1) & 0.003 & 0.001 & 0.003 & 0.000 & 0.005 \\
        \hline
    \end{tabular}
    \caption{Empirical total elementwise correlation, defined in \eqref{eq:ass-neg-tot-corr}, observed in the different datasets. The base models are: lasso regression, ridge regression, linear model, random forest and two different neural nets}
    \label{tab:datasets2}
\end{table}

\begin{table}[h]
    \centering
    \begin{tabular}{c|rrrrr}
    \hline
        & Crime & \makecell{Aquatic\\ toxiticity} & Diamonds & Cement & Productivity\\
        \hline
        COMA (Adapt. $\eta$) & 0.006 & -0.010 & 0.001 & -0.006 & -0.001\\
        COMA ($\eta$=1) & 0.002 & -0.007 & 0.003 & -0.012 & -0.002\\
        \hline
    \end{tabular}
    \caption{Empirical total elementwise correlation, defined in \eqref{eq:ass-neg-tot-corr}, observed in the different datasets. The base models are three random forests and three neural nets with different tuning parameters}
    \label{tab:datasets4}
\end{table}

\subsection{Additional simulations on the negative correlation assumption}
In this section, we investigate the assumption of total negative correlation in Theorem~\ref{prop:cov} via a simulation study. We study the specific case introduced in \citet{chernozhukov2018exact}, where the data is simulated as follows:
\[
y^{(t)} = \beta^\mathrm{T} x^{(t)} + \varepsilon^{(t)}, \quad t = 1, \dots, 200,
\]
where $x^{(t)}$ is distributed as $\mathcal{N}(0, I_{100})$. The serial dependence is induced by generating the error as follows:
\[
\varepsilon^{(t)} = \rho \varepsilon^{(t-1)} + \xi^{(t)}, \quad \xi^{(t)} \stackrel{iid}{\sim} \mathcal{N}(0, 1-\rho^2),
\]
where $\rho = 0.1$ and $\varepsilon^{(0)}=0$. The coefficients are $\beta=(4/\sqrt{5}, 4/\sqrt{5}, 4/\sqrt{5}, 4/\sqrt{5}, 4/\sqrt{5}, 0, \dots, 0)$. At each time step $t=100,\dots,200$, the model is re-trained and a prediction interval for $y^{(t)}$ based on $x^{(t)}$ is constructed using Algorithm 1 in \citet{chernozhukov2018exact} with $\alpha=0.1$ and absolute residuals as score function. In particular, the algorithm chosen in this case is Lasso with 4 different penalty parameters. Under suitable conditions, \citet{chernozhukov2018exact} proves that the coverage is approximately valid.

\begin{figure}[H]
    \centering
    \includegraphics[width = 1\textwidth]{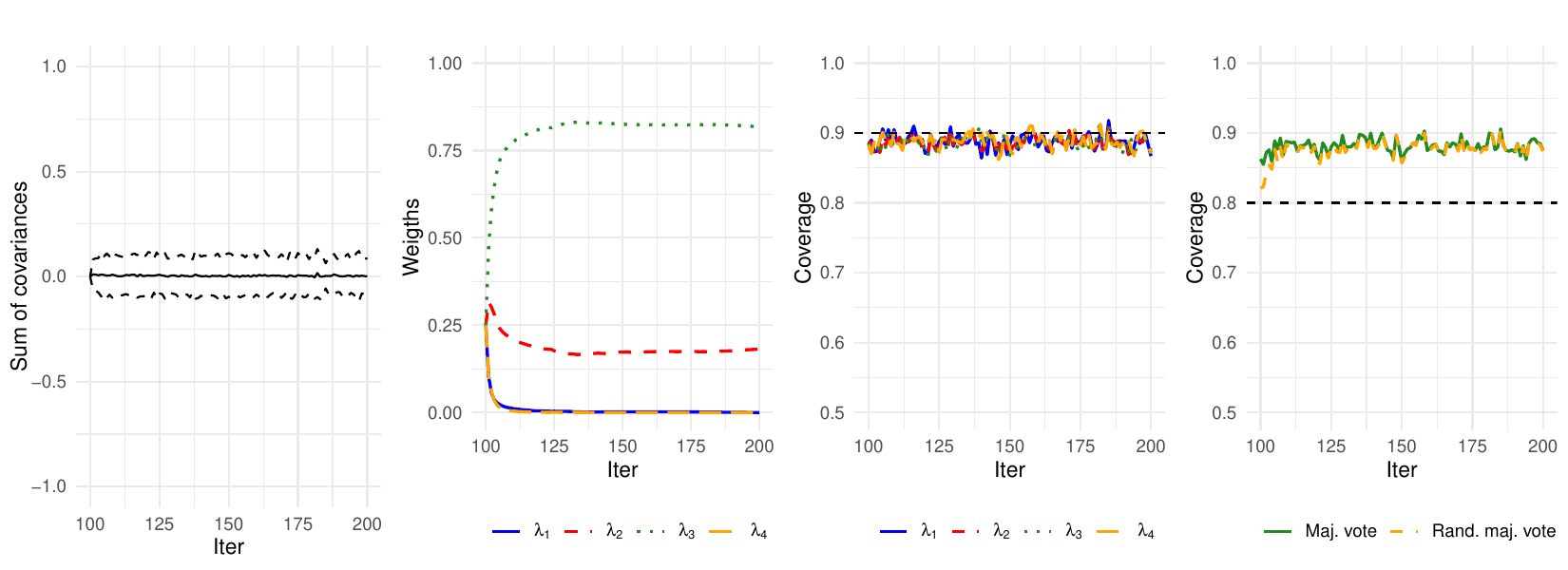}
    \caption{In the first plot, the empirical counterpart of the quantity $\Ev[\sum_{k=1}^K(\phi_k^{(t)} - \Ev[\phi_k^{(t)}])(W_k^{(t)} - \Ev[W_k^{(t)}])]$ is reported ($\pm \mathrm{sd}$), for all $t=100, \dots, 200$. In the second plot, the weights assumed by the various algorithms are reported. The last two plots represent the coverage of the different algorithms (third column) and the coverage of the COMA method with and without randomization (fourth column). The black dashed lines represent the levels $1-\alpha$ and $1-2\alpha$.}
    \label{fig:sim_cov}
\end{figure}

The procedure is repeated $1000$ times, and we report the empirical average over the number of replications. As can be seen in Figure~\ref{fig:sim_cov}, the empirical counterpart of $\Ev[\sum_{k=1}^K(\phi_k^{(t)} - \Ev[\phi_k^{(t)}])(W_k^{(t)} - \Ev[W_k^{(t)}])]$, obtained as the empirical mean of the quantities across the replications, is around 0 for all iterations $t=100, \dots, 200$. Although it is positive in some cases, it is only slightly above zero. The weights for each replication are obtained using the COMA method with an adaptive $\eta$, and on average they concentrate their mass in the penalty parameters $\lambda_2$ and $\lambda_3$. The coverage level for the COMA method is between levels $1-2\alpha$ and $1-\alpha$.

\subsection{Amazon stock prices}
We replicate the analysis described in Section~\ref{sec:google} using the time series of Amazon stock prices from January 1, 2006, to December 31, 2014.

\begin{figure}[H]
    \centering
    \includegraphics[width=\linewidth]{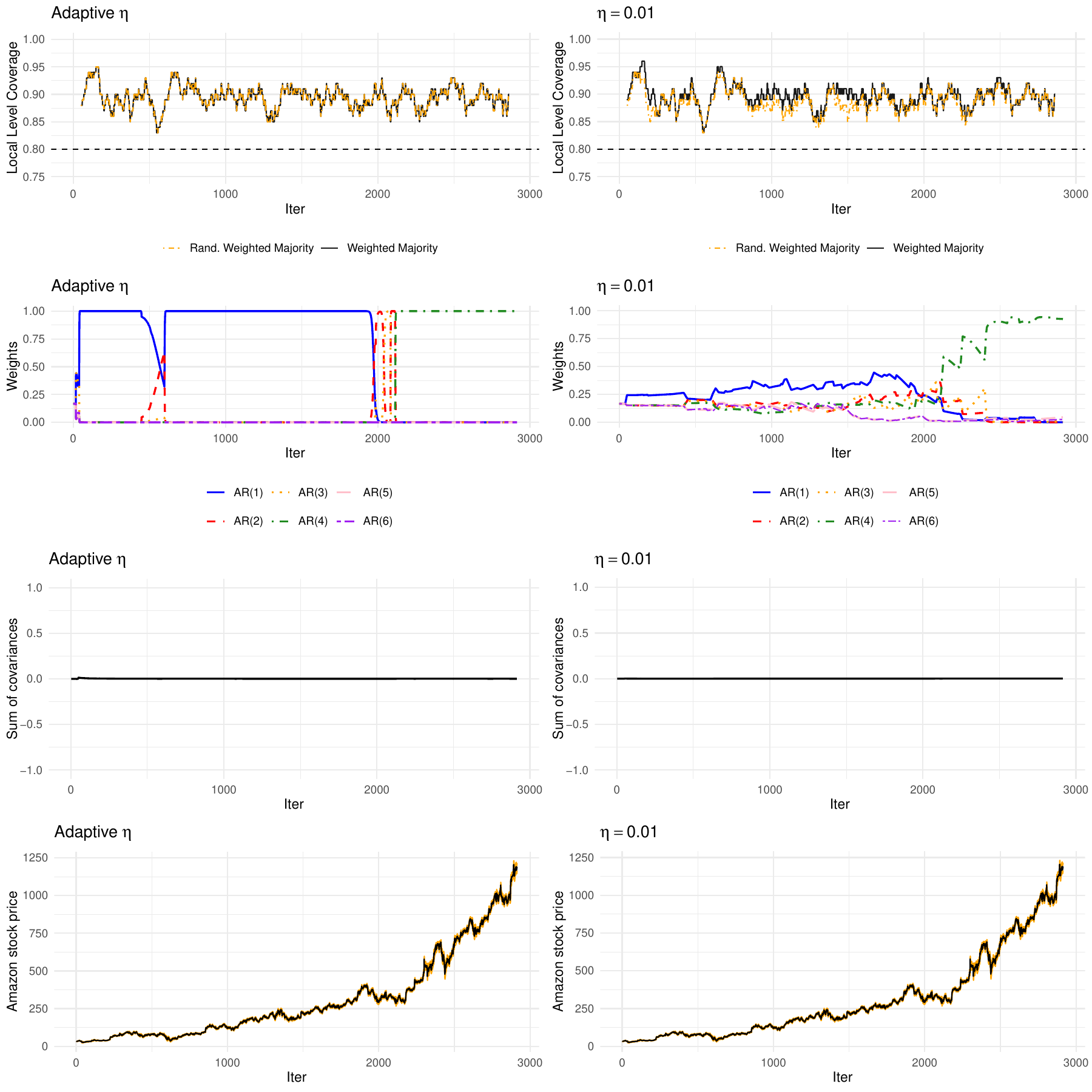}
    \caption{Local level coverage obtained in a window containing 100 data points (first row), weights assumed during the different iterations by the various models (second row), empirical sum of the quantity $\frac{1}{T}\sum_{t=1}^T\sum_{k=1}^K(\phi_k^{(t)}-\bar{\phi}_k)(w_k^{(t)}-\bar{w}_k)$ obtained using an increasing number of observations (third row) and series of the original stock prices with corresponding prediction intervals (fourth row). The black dashed lines in the first row represent the level $1-2\alpha$. The results corresponding to the strategy with adaptive $\eta$ are presented in the first column, while the results corresponding to the strategy with $\eta=0.01$ are presented in the second column}
    \label{fig:amazon}
\end{figure}

The results, reported in Figure~\ref{fig:amazon}, show that the empirical coverage remains close to the level $1-\alpha$. In addition, the weights fluctuate across replications, illustrating that different methods yield varying results in different periods. The total covariance appears to be stable around zero across all iterations.
\end{document}